# SECNLP: A Survey of Embeddings in Clinical Natural Language Processing

Kalyan KS, S. Sangeetha

Text Analytics and Natural Language Processing Lab

Department of Computer Applications

National Institute of Technology Tiruchirappalli, INDIA.

kalyan.ks@yahoo.com, sangeetha@nitt.edu

**ABSTRACT**

Traditional representations like Bag of words are high dimensional, sparse and ignore the order as well as syntactic and semantic information. Distributed vector representations or embeddings map variable length text to dense fixed length vectors as well as capture prior knowledge which can transferred to downstream tasks. Even though embedding has become de facto standard for representations in deep learning based NLP tasks in both general and clinical domains, there is no survey paper which presents a detailed review of embeddings in Clinical Natural Language Processing. In this survey paper, we discuss various medical corpora and their characteristics, medical codes and present a brief overview as well as comparison of popular embeddings models. We classify clinical embeddings into nine types and discuss each embedding type in detail. We discuss various evaluation methods followed by possible solutions to various challenges in clinical embeddings. Finally, we conclude with some of the future directions which will advance the research in clinical embeddings.

## 1. INTRODUCTION

Distributed vector representation or embedding is one of the recent as well as prominent addition to modern natural language processing. Embedding has gained lot of attention and has become a part of NLP researcher's toolkit. According to distributed hypothesis [71, 72], semantic similarity between two words can be expressed in terms of their contexts (i.e.) words with similar contexts have similar meaning. Word embeddings which are based on distributed hypothesis, represent words as dense, low dimensional and fixed length vectors in a continuous vector space and make sure that words with similar meaning are nearby. A word embedding typically consists of hundreds of dimensions and each dimension represents a feature. So, in word embedding the meaning of a word is distributed across dimensions. Word embedding is particularly suitable for deep learning models which consists of multiple layers employing matrix operations to find the high level representations of text data.

For the past few decades, researchers employed machine learning approaches with traditional representations like bag-of-words for NLP tasks. In one hot representation of words, a vocabulary of all the unique words in the corpus is generated and each word is represented as vector of 0s and 1s where the dimension corresponding to the word is set to 1 and all others set to 0. Bag of words feature vector of a document is the sum of one hot vectors of all the words and has 1 only for the words occurring in the document. For a better representation, 0s and 1s can be replaced with other measures like word frequency, tf-idf measure, N-grams etc. All these representations are high dimensional, sparse and ignore order as well as syntactic and semantic similarities of the words.

In contrast, embedding maps the variable length text to dense vector representations and overcome curse of dimensionality and lack of syntactic and semantic information in representations. Moreover, embeddings are learned in an unsupervised manner which capture the knowledge in a large unlabeled corpus and it can be transferred to the downstream tasks with small labeled data sets. Hence, embedding has become an unavoidable choice for text representation in the recent times of deep learning era.

**1.1 Literature Selection**

We collected papers from various sources like PubMed, Google Scholar, ScienceDirect, ACL Web Anthology, and AAAI .We confined to the papers which are published in the period January 2014 to Nov 2018 because of the recent popularity of embeddings. We used keywords like "deep learning," "medical", "clinical," "embeddings", "natural language processing", "distributed representations" and "health" to retrieve the relevant papers and gathered 230 articles. After the removal of duplicate articles as well as the articles which are not related to clinical natural language processing, the number of articles reduced to 120. Finally, we included the most relevant 80 papers after a manual review of all the remaining articles.

Even though embeddings have become de facto standard for text representations for deep learning based NLP tasks in both general and medical domains, there is no survey paper which presents a detailed review of embeddings in the form of classification of embeddings as well as the challenges to be solved. To the best of our best knowledge, we are the first to present a detailed review of embeddings in Clinical Natural Language Processing. Our survey includes

- Classification as well as comparison of Medical Corpora (Section 2)
- Overview of various Medical Codes (Section 3)
- Overview as well as comparison of various Embedding Models (Section 4)
- Classification of Embeddings in Clinical Natural Language Processing (Section 5)
- Evaluation Methods of Embeddings in Clinical Natural Language Processing (Section 6)
- Discussion (Section 7)
- Conclusion and Future Work (Section 8)

**2. MEDICAL CORPORA**

In this section, we classify medical corpora into four types as shown in figure 1 and then discuss each type followed by a comparison (Table I)

Embeddings are inferred using any of the embeddings models over a large unlabeled corpus. Quality of embeddings inferred, depends on two properties of corpus like size and whether it is general or domain specific. A large corpus provides better vocabulary coverage while a domain related corpus provides better semantic representation of terms. Medical corpora can be classified into four categories.

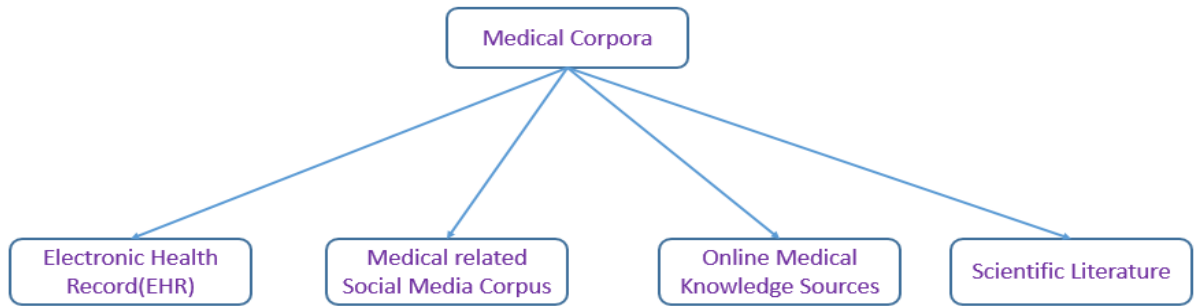

Figure 1. Classification of Medical Corpora

**2.1 Electronic Health Record (EHR)**

In the recent times, Electronic Health Records has become the first option to store patient details in the most of the hospitals [73]. EHRs include both structured data like diagnostic codes, procedure codes, medication codes, laboratory results etc. as well as unstructured data like clinical notes written by health professionals [74]. EHRs containing rich clinical information have become an invaluable source of data for many clinical informatics applications [75 and76]. Some of the research studies have used publicly available EHR data while others have used private EHR data. MIMIC Dataset [77 and 78] is the largest publicly available EHR dataset and is described below

**2.1.1 MIMIC Dataset**

Multiparameter Intelligent Monitoring in Intensive Care (MIMIC) [77 and 78] is a publicly available ICU dataset developed by the MIT Lab. It includes demographics, vital signs, laboratory tests, medications, and more. MIMIC-II [77] contains data collected from Intensive Care Units of Beth Israel Deaconess Medical Center from 2001 to 2008 while MIMIC III [78] consists of data collected in between 2001 and 2012. The data in MIMIC datasets is de-identified and can be used for research purpose. But prior to access, agreement to data use and completion of a training course is mandatory.

**2.2 Medical related Social Media Corpus**

In the recent times, social media evolved as a medium of expression for internet users. Medical related Social Media corpus includes tweets posted by individuals, questions and answers in discussion forums related to health issues. In Twitter[1], users express health related concerns in short text of 140 characters while health discussion forums consists of health related questions raised and the corresponding answers. Some of the popular health discussion forums are MedHelp[2], DailyStrength[3], AskAPatient[4] and WebMD[5].

-----------------------------------

[1]https://twitter.com
[2]https://www.medhelp.org
[3]https://www.dailystrength.org/
[4]https://www.askapatient.com/
[5]https://www.webmd.com/

Social media text is highly informal and conversational in nature with lot of misspelled words, irregular grammar, non-standard abbreviations and slang words. Moreover, the users describe their experiences in non-standard and descriptive words. Analysis of medical social media text which contains rich medical information can provide new medical insights and improved health care.

**2.3 Online Medical Knowledge Sources**

Online medical knowledge sources contain medicine and health related information which is created and maintained by medical professionals. Merriam-Webster Medical Thesaurus[6], Merriam-Webster Medical Dictionary[7] and Merck Manual[8] are some of the online medical knowledge sources. Merriam-Webster Medical Thesaurus consists of word definition along with example sentence, synonyms, related words and antonyms while Merriam-Webster Medical Dictionary consists of word definition along with multiple example sentences and synonyms. Merck Manual which is a medical text book consists of articles related to various topics including disorders, drugs and tests. From these sources, corpus can be built and adopted by any embedding model to generate the embeddings of variable length text sequences like words, phrases and sentences in the corpus.

eMedicine[9] is an online website which consists of almost 6,800 (by December 2018) articles related to various topics in medicine like Emergency medicine , Internal medicine etc. Each article is authored by a certified specialist in the concerned area which undergoes four levels of peer view which includes review by Doctor of Pharmacy. Medical Subject Headings (MeSH)[10] is created and maintained by United States National Library of Medicine[11]. It is a controlled vocabulary used for indexing articles in PubMed and classifying diseases in clinicaltrials.gov

MedlinePlus[12] maintained by United States National Library of Medicine offers reliable and updated information on various topics related to health in an easy to understand language. It is a medical encyclopedia that has information over 1000 diseases and conditions. Sciencedaily[13] and Medscape[14] are two other online sources that provides latest news related to medicine.

**2.4 Scientific Literature**

PubMed[15] maintained by United States National Library of Medicine, is a search engine for citations and abstracts of research articles published in the areas of life sciences and biomedicine. As of December 2018, PubMed has 14.2 million articles with links to full-text. Apart from this, it provides access to books with full text available. PubMed Central (PMC)[16] is a digital repository of research papers published in the areas of biomedicine and life sciences and it provides free access. As of December 2018, it has over 5.2 million articles. Table I gives a comparison of various medical corpora.

------------------------------

[6]https://www.merriam-webster.com/thesaurus
[7]https://www.merriam-webster.com/medical
[8]https://www.msdmanuals.com/
[9]https://emedicine.medscape.com/
[10]https://www.ncbi.nlm.nih.gov/mesh
[11]https://www.nlm.nih.gov/
[12]https://medlineplus.gov/
[13]https://www.sciencedaily.com/
[14]https://www.medscape.com/
[15]https://www.ncbi.nlm.nih.gov/pubmed/

**TABLE I**

**COMPARISION OF MEDICAL CORPORA**

| S.No. | Medical Corpus | Contains | Language Used | Noisy | Access | Authors | Example |
|---|---|---|---|---|---|---|---|
| 1 | Electronic Medical Records (EMR) | Patient information in the form of medical codes, laboratory results and clinical notes | Professional | Yes, with lot of unstandardized abbreviations and misspelled words in clinical notes | Restricted access because of sensitive information | Trained medical professionals in hospitals | MIMIC II[77] and MIMIC III[78] |
| 2 | Medical Social Media | i) Views and opinions related to health in the form of tweets ii) Health related questions and answers in Health Discussion Forums | Colloquial | Yes, with irregular grammar, slang words and misspelled words | Free Access | Common Public | Twitter[1], AskAPatient[4], WebMD[5], MedHelp[2] and DailyStrength[3] |
| 3 | Online Medical Knowledge Sources | i) Medical words definitions, synonyms and related words ii) Medical Articles iii) Latest Medical News | Professional | No | Free Access | Trained Medical Professionals | Merriam Webster Dictionary[6] and Thesaurus[7], eMedicine[9], Medscape[14], MedLinePlus[12], Sciencedaily[13], MeSH[10] and MerckManuals[8] |
| 4 | Scientific Literatures | Abstracts, Citations as well as full text of life sciences and biomedical research articles | Professional | No | Free Access | Researchers in life sciences and biomedical areas | PubMed[15] and PubMed Central[16] |

3. **MEDICAL CODES**

The primary motive behind EHR [70] is to record the patient information right from admission to discharge in a systematic way. Several classification schemes are available for recording the relevant clinical information. For example, ICD (International Statistical Classification of Diseases and Related Health Problems) - diagnosis codes, CPT (Current Procedural Terminology) - procedure codes, LOINC (Logical Observation Identifiers Names and Codes) - laboratory codes, and RxNorm- medical codes. Table II gives brief summary of the medical codes.

------------

[16]https://www.ncbi.nlm.nih.gov/pmc/

**TABLE II**

**SUMMARY OF MEDICAL CODES**

| Schema | Description | Number of Codes | Examples |
|---|---|---|---|
| ICD-10 (Diagnosis) | Prepared by World Health Organization(WHO) and contains codes for disease, signs and symptoms etc. | 68,000 | *'R070'* : Pain in Throat<br>*'H612'* : Impacted cerumen |
| CPT(Procedures) | Prepared by American Medical Association(AMA) and contain codes for medical, surgical and diagnostic services | 9,641 | *'90658'* :Flue Shot<br>*'90716'* : Chicken Pox Vaccine |
| LOINIC (Laboratory) | Prepared by Regenstrief Institute, a US nonprofit medical research organization and contain codes for laboratory observations | 80,868 | *'8310-5'* : Body Temperature<br>*'5792-7'* : Glucose |
| RxNorm (Medications) | Prepared by US National Library of Medicine and is a part of UMLS. Contains codes for all the medications available in US market. | 1,16,075 | *'1191'* : Aspirin<br>*'215256'* : Anacin |

These standardized codes are used to ensure consistency in recording patient information and other applications like reimbursement claims. Most of the US health care payment systems are based on these standardized codes. As an example, the health care insurance companies pay the reimbursements based on the medical codes assigned to the clinical reports [28].

4. **EMBEDDING MODELS**

This section gives a brief overview as well as comparison of various embedding models like word2vec (section 4.1), paragraph2vec (Section 4.2), glove (Section 4.3), fasttext (Section 4.4) and Elmo (Section 4.5). Table IV gives a summary of various embedding models and Table V gives a comparison of various embedding models.

Embedding is one of the promising applications of unsupervised learning as well as transfer learning because embeddings are induced from large unlabeled corpora and the prior knowledge captured in embeddings can be transferred to downstream tasks involving small datasets. Embedding models can be classified into Prediction based and Count based [129]. Prediction based models learn embeddings by predicting target word based on context words or vice versa. Count based models learn embeddings by leveraging global information such as word context co-occurrences in a corpus.

Research in learning distributed vector representations started with Rumelhart et al. [79]. Several research studies [80]-[86] laid foundation for research in embeddings. Bengio et al. [82] proposed a neural network based model for the task of next word prediction. The model consists of hidden layer with *tanh* activation and output layer with *softmax* activation. The output layer produces the probability of all the words in the vocabulary for the given *n-1* input words. In doing so, the model learns distributed representations of the words. The model overcomes the problems of curse of dimensionality and unseen sentences.

Collobert and Weston [85] were the first to demonstrate the usage of pre trained word embeddings. They proposed CNN based model which takes a sentence as input and outputs parts of speech tags, named entity tags, chunks etc. Finally, the model proposed by Mikolov et al. [62] gained lot of attention, brought immense popularity and made embeddings the first choice for text representation. Later, Glove [87] proposed by Pennington et al., FastText [88] proposed by Bojanowski et al. and ELMo [123] proposed by Peters et al. have become other popular embedding models.

4.1 **Word2Vec**

Inspired by distributed hypothesis [71, 72] and neural language models, Mikolov et al. [62] proposed word2vec, a simple and efficient algorithm for inferring dense vector representations of words from a large unlabeled corpus. It is a shallow neural network model that learns the word representations by optimizing the objective function which involves both target word and context word. Word2vec builds vocabulary out of the corpus and learns the word representations by training a three layered neural network. Word2vec offers two models namely Continuous Bag of Words (CBOW) and Skip-gram. CBOW learns the representations by predicting the target word based on its context words while skipgram learns representations by predicting each of the context words based on the target word. So, one has to choose one of the architectures and set values for hyper parameters like embedding size, context size , minimum frequency for a word to be included in the word vocabulary to generate the word embeddings from a large corpus of unlabeled data.  Table III gives a summary of various hyper parameters in word2vec model.

**TABLE III**

**SUMMARY OF HYPER PARAMETERS IN WORD2VEC**

| Parameter | Default Value | Meaning |
| --- | --- | --- |
| size | 100 | Dimension of vectors |
| window | 5 | Size of context window |
| min_count | 5 | Minimum frequency of a word to be included in the vocabulary |
| workers | 3 | Number of threads to train the model |
| sg | 0 | 0 means CBOW model is used and 1 means skipgram is used. |
| hs | 0 | <ul><li>1 for hierarchical softmax</li><li>0 and 'negative' with non-zero value means, negative sampling is used</li></ul> |
| negative | 5 | <ul><li>0 means, no negative sampling</li><li>>0 means negative sampling is applied and the value represents the number of noise words to be used.</li></ul> |

In Word2vec, there are two options to evaluate the generated representations namely, distance and analogy. Distance option allow to retrieve the most semantically similar words for a given word in the corpus along with the cosine similarity score. In this context, cosine similarity score represents the degree of semantic similarity between the two words.  Analogy option allows to find the linguistic regularities like 'king-man+women=?'

### 4.1.1 Continuous Bag of Words (CBOW)

CBOW model learns embeddings by predicting the target word against its context words. CBOW model can be viewed as supervised model with the context words as input and the target word as output.

For example, consider the sentence, "the black pen is on the red table". With a context window of size 2, the (context, target words) pairs are ([black, pen], the), ([the, pen, is], black), ([the, black, is, on], pen) and so on.

As in figure 2, CBOW model consists of three layers namely input layer, hidden layer and output layer. The layers are connected by two weight matrices W and W'. The input layer takes the one hot vectors of context words as input and the output layer applying softmax function predicts the one hot vector of target word. Error between the original and predicted vectors is back propagated to update the weight matrices W as well as W'. Finally for each word in the vocabulary of given corpus, two vectors $V_c$ and $V_w$ are obtained (i.e., $V_c$ is from W and $V_w$ is from W'). The objective function of CBOW model is

$$J = \frac{1}{V} \sum_{i=1}^{V} \log p(w_i \mid w_{i-n}, \ldots w_{i-2}, w_{i-1}, w_{i+1}, w_{i+2}, \ldots w_{i+n})$$

where V is the size of vocabulary and n is the window size.

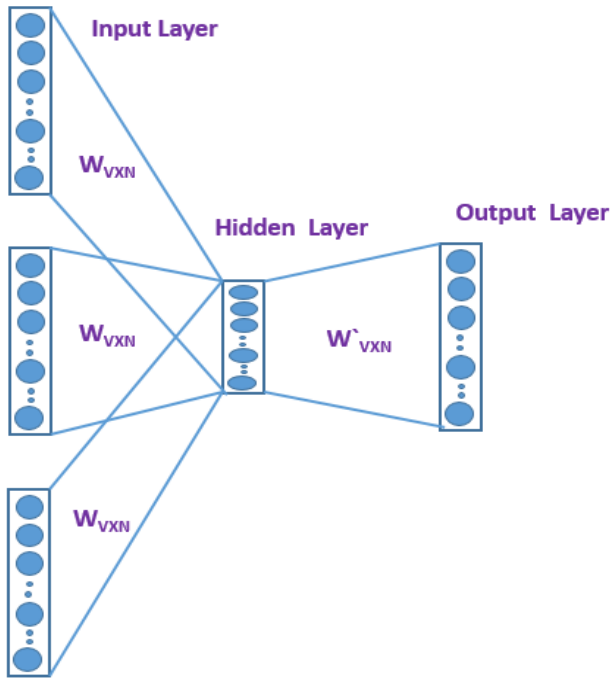
**Figure 2. CBOW Model [110]**

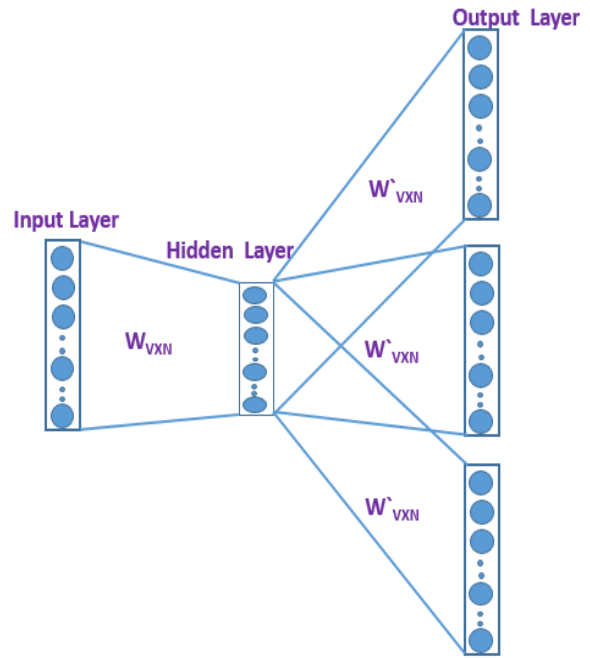
**Figure 3. Skipgram Model [110]**

**4.1.2 Skipgram**

As in figure 3, Skip-gram model works exactly opposite to CBOW and learns the embeddings by predicting the context words against the target word. Skipgram model can be viewed as supervised model with the target word as input and the context words as output.

> For example, consider the sentence, "the black pen is on the red table". With a context window of size 2, the (target_word, context) pairs are (the, [black, pen]), (black, [the, pen, is]), (pen, [the, black, is, on]) and so on.

The objective function of skipgram model is

$$J = \frac{1}{V} \sum_{i=1}^{V} \sum_{-n \leq j \leq n, \neq 0} \log p(w_{i+j} \mid w_j)$$

where V is the size of vocabulary and n is window size.

Each of CBOW and Skipgram models have their own advantages and disadvantages. In both the models, learning output vectors is computationally expensive. To address this issue, two methods namely Negative sampling and Hierarchical softmax is proposed. Negative sampling limits the number of output vectors to be updated while Hierarchical softmax is based on Huffman tree. Negative sampling works well with in-frequent words, whereas Hierarchical softmax works well with frequent words [130].

**4.2 Doc2Vec**

Paragraph2vector popularly known as Doc2vec, is an extension to Word2vec [62] and is proposed by Le et al. [89]. It is an unsupervised model which maps the variable length text like sentences, paragraphs and documents to dense vector representations. Doc2vec learns the dense vectors representations for both variable length text and the words in the text. It offers two models namely Distributed Bag of Words (DBOW) and Distributed Memory (DM).

4.2.1 **Distributed Memory (DM)**

DM model is similar to Continuous Bag of Words model of Word2vec [62]. CBOW predicts the center word from the context words while DM predicts the next word using the concatenation or average of the vectors of paragraph and context words. As in figure 4, it consists of three layers. First layer takes the vectors of paragraph and context words as input. Second layer concatenates or average both these vectors. The final layer which is a classifier, predicts the vector for the next word.

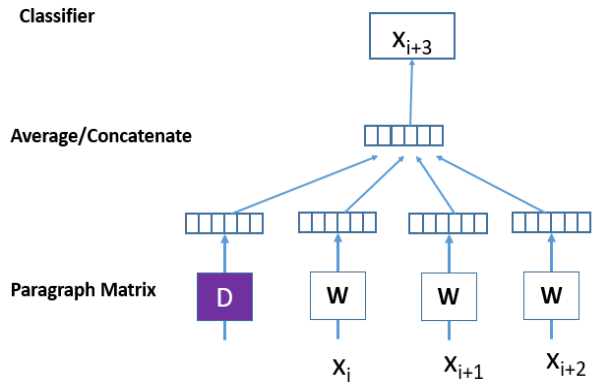 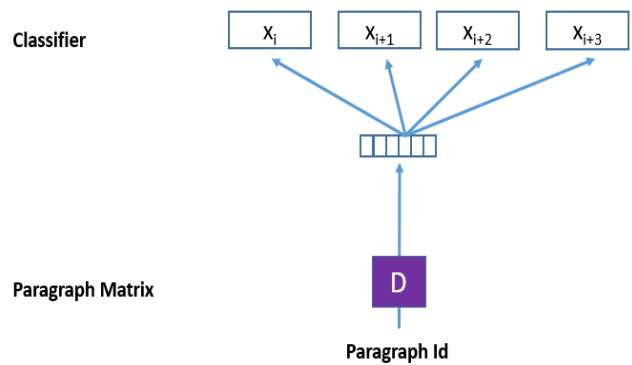

**Figure 4. DM Model [89]**          **Figure 5. DBOW Model [89]**

#### 4.2.2 Distributed Bag of Words (DBOW)

DBOW model is similar to Skipgram model of Word2vec [62]. Skipgram predicts the context words from the center word while DBOW predicts the context words using the paragraph. As in figure 5, it consists of three layers. First layer takes the vector of paragraph as input. Second layer is the hidden layer. The final layer which is a classifier, predicts the vectors of the context words.

In both the DBOW and DM models, the matrix D has dense vector representations. Each column of D is an embedding of the variable length text. Compared to DM model, DBOW model is simple, needs less memory. DM model stores softmax weights as well as word vectors while DBOW model stores only softmax weights. According to Le et al., [89] DM works well for most of the tasks but recommended to use a combination of vectors from DM and DBOW, as the combination gives consistent results across tasks.

### 4.3 Glove

Global Vectors for word representations popularly knowns as "Glove" [87] is proposed by Pennington et al. Methods like LSA which uses matrix factorization utilize global co-occurrence statistics but perform poorly in word analogy task. While methods like Word2vec [62] do well in word analogy tasks but poorly utilize the global co-occurrence statistics. The Glove model combines the advantages of Word2vec model in learning representations based on context as well as matrix factorization methods in leveraging the global co-occurrence statistics.

In Glove, a word occurrence matrix is generated in which rows represent the words and columns represent the context. Each value in the matrix represent how frequently a word co-occurs with a context. Factorization of the word co-occurrence matrix results in a low dimensional matrix where rows represent words and columns represent features. Each row in low dimensional word-feature matrix represents the dense vector representation of word where the size of feature can be preset to the required value. The objective function in Glove model is

$$J = \sum_{i,j=1}^{V} f(X_{ij}) \, (w_i^T \widetilde{w}_j + b_i + \tilde{b}_j - \log X_{ij})^2$$

Here f is a weighting function and $w_i$, $w_j$, $b_i$ and $b_j$ are the learned where $w_i$ and $w_j$ are vectors for main and context words, bi and $b_j$ are biases for the main and context words.

Glove and word2vec has a similarity. Both the models learns the representations based on the contexts. But, Word2vec leverage local contexts whereas Glove utilize the global contexts in the form of global co-occurrence statistics.

**4.4 FastText**

FastText embedding [88] as well as FastText classifier [90] is proposed by Facebook researchers in 2016. FastText embedding is essentially an extension and improvement of the word2vec model [62] proposed by Mikolov et al. In models like word2vec and glove, words are treated as atomic entities and mapped to dense vector representations. So, these models ignore morphological information which is a concern for languages with large vocabulary and many rare words. In FastText model, each word is treated as a bag of character n-grams. Each character n-gram is mapped to dense vector and the sum of these dense vectors represent the word.

> For example, the vector for the word 'mango' is sum of the vectors of '<man', 'mang', 'ango', 'ngo>', mango, 'ango>' etc. (assuming values of 4 and 5 for the hyper parameters *minn* and *maxn* which represents the sizes of smallest and largest character ngrams)

FastText by leveraging the sub word information, offer better representations for rare words because even if the words are rare, their character ngrams appear across the words in the corpus. Further, representations for out of vocabulary words can be obtained using the sum of character ngrams.

4.5 **ELMo**

Embeddings models like Word2Vec[62], Glove[87], FastText[88] assign a single vector representation to a word independent of the context in which it is used. However, meaning of a word changes according to the context in which it is used. In the recent times, a number of models like CoVE [125], TagLM[128], Context2vec[127] have been proposed to generate context dependent representations. However these models have some drawbacks. CoVE needs labeled data to generate context dependent representations and use zero vectors for OOV words. Further CoVE, TagLM and Context2vec models make use of only the last layer representations.

Peters et al. [123] proposed ELMo (figure 6) which generates embeddings for words considering its context, by making use of Character embeddings and BiLSTM. ELMo model consists of L-layers of BiLSTMs with character level word embeddings as input. Contextualized embedding of each word is obtained as a weighted sum of character level word embeddings and the hidden states of BiLSTM.

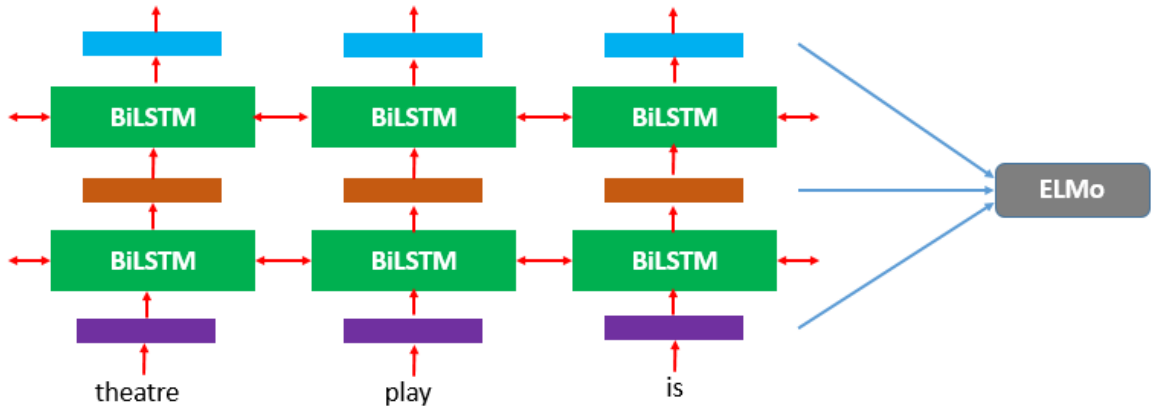

**Figure 6.** ELMo Model [123]

TABLE IV

SUMMARY OF EMBEDDING MODELS

| Model | Architecture | Advantages | Disadvantages |
|---|---|---|---|
| CBOW [62] | Log Bilinear | • Faster compared to skipgram model.<br>• Represents frequent words well. | • Ignore morphological information as well as polysemy nature of words.<br>• No embeddings for OOV, misspelled and rare words. |
| Skipgram [62] | Log Bilinear | • Efficient with small training datasets.<br>• Represents infrequent words well. | • Ignore morphological information as well as polysemy nature of words.<br>• No embeddings for OOV, misspelled and rare words. |
| PV-DM [89] | Log Bilinear | • PV-DM alone give good results for many of the tasks. | • Compared to PV-DBOW, requires more memory as it is needed to store Softmax weights and word vectors. |
| PV-DBOW [89] | Log Bilinear | • Need to store only the word vectors and so requires less memory.<br>• Compared to PV-DM, it is simple and faster. | • Need to be used along with PV-DM to give consistent results across tasks. |
| Glove [87] | Log Bilinear | • Combines the advantages of word2vec model in learning representations based on context as well as matrix factorization methods in leveraging global co-occurrence statistics. | • Ignore morphological information as well as polysemy nature of words.<br>• No embeddings for OOV, misspelled and rare words. |
| FastText [88] | Log Bilinear | • Encode morphological information in word vectors.<br>• Embeddings for OOV, misspelled and rare words.<br>• Pretrained word vectors for 157 languages. | • Computationally intensive and memory requirements increases with the size of corpus.<br>• Ignore the polysemy nature of the words. |
| ELMo [123] | BiLSTM | • Generate context dependent vector representations and hence account for polysemy nature of words.<br>• Embeddings for OOV, misspelled and rare words. | • Computationally intensive and hence require more training time. |

$$ELMo_k^{task} = r^{task} \sum_{j=0}^{L} s_j^{task} h_{kj}^{LM}$$

Here j and k are the indices of the word and the BiLSTM layers, $r^{task}$ is a task specific scaling factor, $s_j^{task}$ represents softmax normalized weights.

## 5. CLASSIFICATION OF MEDICAL EMBEDDINGS

As shown in figure 7, medical embeddings can be classified into nine categories with each category mapping variable length text (e.g. character, words, phrases, sentences or documents) or medical codes or CUIs to dense vector representations. In this section, we discuss about various types of clinical embeddings. Table VI contains a summary of various clinical embeddings and Table VII contains a summary of various clinical NLP tasks with embeddings as input features.

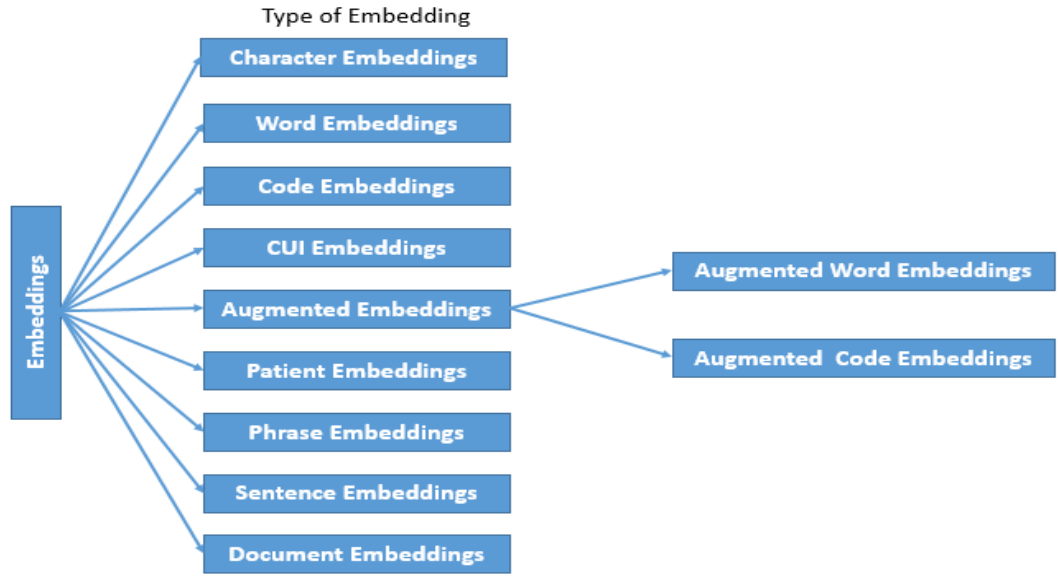

**Figure 7.** Classification of Clinical Embeddings

### 5.1 Character Embeddings

Character embedding models consider character as an atomic unit and maps it to a fixed length dense vector. Vector representation for a word is obtained from the embeddings of the constituting characters using CNN or LSTM. As a result, character embedding models can generate embeddings for Out of Vocabulary (OOV) words, rare words and misspelled words.

Niu et al. [111] proposed deep learning framework based on multi-task character level attention mechanism to map medical concept mention to standard medical vocabulary. Character level embeddings avoids the OOV issue while attention mechanism exploits the morphological information by generating the attention weights on domain related positions which aids the concept normalization. The input concept is encoded into character embedding matrix and it is added with the attention weights. The augmented character embedding matrix is taken as input by character level CNN and then assign the corresponding label. Evaluation on three real world datasets showed the effectiveness of the model.

**Table V**

**COMPARISON OF EMBEDDING MODELS**

| Model | Type | Generate Embeddings for | Vectors for OOV Words | Encode Morphological Information | Use of global co-occurrence statistics | Context Independent / Context Dependent Word Vectors |
|---|---|---|---|---|---|---|
| CBOW | Prediction based | Words | No | No | No | Context Independent |
| Skipgram | Prediction based | Words | No | No | No | Context Independent |
| PV-DM | Prediction based | Sentences, Paragraphs and Documents | - | No | No | - |
| PV-DBOW | Prediction based | Sentences, Paragraphs and Documents | - | No | No | - |
| Glove | Count based | Words | No | No | Yes | Context Independent |
| FastText | Prediction based | Characters N-grams and Words | Sum of character n-grams in the word | Yes | No | Context Independent |
| ELMo | Prediction based | Words | Generated over character embeddings using CNN or BiLSTM | Yes | No | Context Dependent |

### 5.1 Word Embeddings

Word embeddings models map words to dense vector representations as well as capture syntactic and semantic information. Context independent word embedding models like word2vec [62], glove [87] and fasttext [88] assign single vector representation for a word ignoring the context in which it appears. Context dependent word embeddings models like Elmo [123] assign different representations for a word depending on the context in which it appears.

Huang et al. [1] experimented with embeddings induced from various corpora in order to evaluate the effectiveness of embeddings in modelling medical terms and relations. Using Word2Vec, generated word embeddings separately over the text collected from health discussion forum, PubMed[15] abstracts and Wikipedia corpus. The text in these corpora vary greatly in terms of style and vocabulary of the language used. Results demonstrated that the model learned over health forum text represent popular medical terms from Ranker[17] significantly better compared to the model learned over PubMed abstracts authored by professionals. Further, the word spaces induced by the health forum text and PubMed abstracts better model the medical terms compared to the word space induced by general purpose Wikipedia Corpus.

----------------

[17] https://www.ranker.com

Lee et al. [4] explored the use of word embeddings generated from various medical knowledge sources using Word2Vec, for the task of medical concept normalization. Medical Concept normalization maps the health condition expressed in lay terms to the standard medical concepts and is treated as multi label classification problem. Evaluation on two standard datasets Twitter [11] and AskAPatient [91] showed that RNN and CNN models trained over the embedding learned from combined clinical data sources outperformed the baselines.

Miftahutdinov et al. [10] proposed novel architectures based on LSTM and GRU with attention to the problem of mapping disease mention in free form social media text to standard UMLS (Unified Medical Language System)[18] concepts. Experimented with two publicly available word embeddings. First one is HealthVec [17]: learned over the corpus consisting of medical related user comments in health websites. Second one is PubMedVec[39]: trained over the PubMed abstracts. Experimental results demonstrate the superior performance of LSTM and GRU based models with attention compared to the baselines based on CNN. Further, showed that the addition of UMLS based semantic similarity features improved the performance.

Limsopatham at al. [11] demonstrated the importance of semantic level information in the task of medical concept normalization with the use of CNN and RNN based neural network models. Experimented with two word embeddings. First one is GNews[19], a publicly available word embeddings generated from Google News Corpus using word2vec and the other is BMC, word embeddings trained on biomedical literature downloaded from BioMed Central[20] using word2vec. Evaluated the models on three datasets having text extracted from Twitter and blog posts. Results showed that CNN with GNews word embeddings gave the best performance across all the three data sets. Further updating the word embeddings (allowing the model to update) improved the performance only on large test data.

Belousov at al. [12] presented an ensemble system based on multinomial log regression and Bi-GRU deep learning model for the mapping of informal mentions in free text obtained from drug related tweets to MedDRA11 codes. Used three pre trained word embeddings namely GNews (trained on Google News Corpus), Twitter [63] (trained on general purpose tweets) and DrugTwitter [18] (trained on drug related tweets). Multinomial log regression trained on all the three embeddings which are combined using mean rule outperformed Bi-GRU trained on GNews. Further, ensemble of both the models based on mean rule achieved the highest accuracy.

--------------------------------

[18] https://www.nlm.nih.gov/research/umls/

[19] https://code.google.com/archive/p/word2vec/

[20] https://www.biomedcentral.com

**TABLE VI**

**SUMMARY OF CLINICAL EMBEDDINGS**

| | Type of Embedding | Key Points | Papers |
|---|---|---|---|
| Embeddings | Char Embeddings | • Include morphological information and can generate embeddings for OOV, misspelled and rare words.<br>• Meaning is not encoded. | [111] |
| | Word Embeddings | • Incorporate syntactic and semantic information but not morphological information.<br>• No embeddings for OOV and misspelled words. | [1, 3, 4, 10, 11, 12, 14-23, 26, 29, 31, 32, 34, 35, 37, 39-41, 49, 51, 60, 61, 112, 114-117, 119, 120, 122, 124 and 126] |
| | Code Embedding | • Finds applications in various tasks in health care analytics.<br>• Ignoring sequential information in codes affects the quality of inferred embeddings. | [8, 9, 43, 45 and 47] |
| | CUI Embeddings | • Use of CUI embeddings includes domain knowledge from the controlled vocabularies.<br>• Finds applications in information retrieval and analytics related to clinical domain. | [9 and 13] |
| | Augmented Embeddings | • Improved quality of embeddings with the addition of domain knowledge.<br>• No rule of thumb regarding the best way of including domain knowledge. | [2, 5, 36, 44, 55 and 56] |
| | Patient Embeddings | • Finds applications in many tasks related to clinical domains.<br>• Ignoring the sequential information in codes can potentially affect the quality of inferred embeddings. | [7, 30, 33, 38, 48, 52, 54, 58 and 59] |
| | Phrase Embeddings | • Phrase embedding can be generated from the aggregation of embeddings of words in the phrases or directly using word2vec or paragraph2vec models. | [113 and 121] |
| | Sentence Embeddings | • Sentence embedding can be generated from the aggregation of embeddings of words in the sentence or directly using paragraph2vec model. | [24, 50 and 57] |
| | Document Embeddings | • Document embedding can be generated from the aggregation of embeddings of sentences or directly using paragraph2vec model. | [25, 42, 46 and 118] |

Minarro-Giménez et al. [14] generated three word embedding models using word2vec from the three corpora (PubMed, Wikipedia and combined) comprising of clinically relevant text obtained from PubMed, Wikipedia, Merck Manuals and Medscape. Evaluated the learned representations in identifying pharmaceuticals properties like relationships to diseases or physiological processes through comparison with National Drug File – Reference

Terminology (NDF-RT)[21] ontology and embeddings induced from combined corpora gave better results.

Sentences in medical text are comparatively short in length and lacks contextual information which results in feature sparseness and semantic ambiguity. Shen et al. [15] proposed cluster based semantic expansion method to include more semantic information and enhance the word embeddings. Generated the word embeddings using skipgram over the corpus consisting of medical records and medical text book. Generated the word clusters using Hierarchical Agglomerative Clustering and then added cluster center vectors to the word embeddings of each word. LSTM with word cluster embeddings outperformed the baselines with just word embeddings in short text classification in both general and medical domain which highlights the importance of adding more semantic information in the form of word embedding cluster centers.

Liang et al. [16] improved the disease text classification with the novel method consisting of rule based features and knowledge based deep learning techniques. Generated word embeddings from MIMIC III [78] dataset using skip gram. Experimented with different word embedding sizes 100, 200, 300, 400, 500 and 600 dimensions and found that 200 dimension embedding gave the best results. Trained the multi view CNN using the pre trained MIMIC III embeddings and the UMLS CUI embeddings [13]. Evaluation on 2008 i2b2 Obesity Challenge [92] demonstrated the superior performance of the proposed model. The results showed that addition of domain knowledge in the form of CUI embeddings improved the performance.

Miftahutdinov et al. [17] proposed a CRF architecture with hand crafted features and word embeddings for the task of identifying disease and drug related expressions in user comments. Treated the problem as sequence labelling and employed CRF which takes a sequence of tokens as input, calculates and assigns the label with highest probability. Experimented with two word embeddings namely HealthVec and PubMed[39]. HealthVec is obtained using CBOW Word2Vec over the corpus having the comments collected from various websites and PubMed is learned over the PubMed biomedical literature. Experimental results show that CRF model with handcrafted features and HealthVec word embeddings outperformed the baselines.

Nikfarjam at al. [18] proposed, a novel architecture called ADRMine based on CRF for extracting adverse drug reaction mentions in social media posts. Experimented with different features which includes lexicon based, contextual, grammatical and semantic. Word clusters based on the word embeddings generated from unlabeled user comments in social media using word2vec gives the semantic features. ADRMine outperformed all the baselines with an F-measure of 0.82. Further analysis showed that contextual and word cluster features significantly contributed for the improved performance.

--------------------------------------------

[21] https://bioportal.bioontology.org/ontologies/NDFRT

Nikhil et al.[19] viewed the problem of inferring the associated medical persona through the lens of multi label classification. Experimented with feature based and neural network based methods on blog and tweets datasets. Tried five different pre trained word embeddings learned over medical corpora as well as generic corpora [93], [87], [39]. For blog data set, averaged word2vec model with Glove Embeddings trained on Web Crawl2 and for tweet dataset CNN model with Glove Embeddings trained on Web Craw2 gave better results. Experimental results showed that diversity of text used for word embedding generation is more important compared to the coverage of medical concepts.

Yuan Li[20] proposed a novel data driven technique based on CNN for disease diagnosis prediction from admission notes. Formulated the problem of disease diagnosis prediction as multi class classification with ten categories. Generated embeddings with a dimension of 128 using FastText on the MIMIC III [78] corpus. Initialized the embedding layer of CNN model with these embeddings for multi class classification. Results demonstrated that CNN model with FastText embeddings trained on MIMIC III outperformed the baselines.

Amitabha [21] viewed the problem of ICD 9 code assignment through multi label multi class classification lens and experimented with different models including CNN with attention. As word embeddings were missing for more than half of domain specific words in Glove pre trained word embeddings, generated word embeddings using Glove algorithm on MIMIC III corpus. Found that CNN with attention gave the best performance in assigning ICD 9 codes to discharge notes.

Limsopatham et al. [22] studied the use of generic word embeddings in domain specific tasks. The author experimented with two approaches that combined the generic and domain embeddings at the input layer and fully connected layer of CNN model for sentence classification. The author used the GNews embedding as generic embedding and generated the domain specific embeddings from two medical corpora using skip gram model from word2vec. For out of vocabulary words, used a vector of random values sampled from [-0.25.0.25]. CNN with word embeddings combined at fully connected layer outperformed all other methods in ADR tweet classification [94] and showed that generic word embeddings provide a good coverage of vocabulary and domain specific word embeddings provide effective semantic representation of domain words.

Miftahutdinov at al.[23] studied the problem of ICD 10 coding of death certificates using LSTM based encoder-decoder model. Formulated the problem of ICD 10 coding as sequence labelling. Experimented with different pre trained word embeddings like HealthVec[17], PubMedVec[39] and GNews. Included prior knowledge in the form of cosine similarity vectors between text and dictionary entry. Used randomly sampled vectors for out of vocabulary words. Evaluation on standard dataset from CLEF eHealth 2017[95] challenge shows that the proposed model with PubMedVec embeddings and prior knowledge outperformed all other methods.

Jagannatha et al. [26] studied various models based on LSTM and CRF with word embeddings as input for sequence labeling in clinical text. LSTM models the long term dependencies and extract the features. With the extracted features as input, CRF calculates the probabilities and assign the label with the highest probability. Word embeddings with a dimension of 200 were inferred using skip-gram model over a combined corpus of PubMed articles, Wikipedia articles and clinical notes from electronic health records. Results showed that LSTM-CRF models with embeddings trained over combined corpora outperformed the baselines.

Jagannatha et al. [27] explored the effectiveness of LSTM and GRU models for medical event detection in electronic health records. Inferred word embeddings with a context window of size 10 and dimension of 200 using skip gram model from a combined corpus of PubMed articles, English Wikipedia and unlabeled EHR notes. Initialized the embedding layers of LSTM and GRU models with these embeddings. LSTMs and GRU models are trained both at sentence and document levels. Both the RNN models outperformed the CRF baselines and in particular GRU model trained at document level gave the best performance.

Sandeep and Olive[28] explored the use of LSTM based architecture in the task of assigning ICD 9 codes to discharge summary notes from MIMIC III. To initialize the word embedding layer, pre trained Glove word embeddings trained over Common Crawl with a dimension of 300 is used. Used zero vectors for words with missing word embeddings. Experimental results showed that the proposed model with pre-trained Glove embeddings achieved considerable improvement in the performance.

Zhang [29] studied Convolutional residual model for assigning disease codes for plain medical text and treated problem as a multi label classification. Convolutional residual model consists of CNN with Deep Residual Network on its top. CNN encodes the plain medical text into fixed length vector. CNN embedding layer is initialized with GoogleNewVec and is further tuned during training. Deep residual network capture label correlation and incorporate information from the encoded sentence vector. Evaluation on a real EHR data demonstrated the effectiveness of the proposed model.

Newman et al. [31] applied LSTM based architecture [96] for extracting the entities related to patient mobility from clinical notes. Explored various domain adaption methods of word embeddings to make up for the small size of medical corpora. Using word2vec and fasttext, generated different embeddings over various corpora which includes in-domain and out-of domain corpora. Experimental results shows that domain adaption of word embeddings improved the performance. Further, the methods like concatenation and pre initialization achieve highest performance, nonlinear mapping method showed consistent performance across experiments.

Yuan Luo [32] explored LSTM based sentence and segment level models for relation classification in clinical notes. Experimented with general domain GoogleNewsVec embeddings as well as the medical domain embeddings generated from MIMIC III corpus using word2vec. Both the embeddings are of dimension 300. Evaluation on i2b2 relation classification dataset [97] showed that segment level LSTM outperformed all the baselines including sentence level LSTM. Further showed that medical domain word embeddings improved the performance compared to general domain embeddings.

Wang et al. [34] studied word embeddings induced from various corpora like clinical notes, biomedical research papers, Wikipedia articles and news articles and provided a comparison. Using skip gram algorithm, inferred embeddings from clinical notes and biomedical literature and used the pretrained Glove and GoogleNews embeddings. Evaluated the embeddings both qualitatively and quantitatively. Experimental results showed that embeddings inferred from clinical notes and biomedical research papers model the medical terms better compared to Glove and GoogleNews embeddings. Further, the embeddings induced from biomedical corpora do not necessarily outperform the embeddings inferred from general corpora in any biomedical NLP application.

Mullenbach et al. [35] proposed CAML(Convolutional Attention for Multi-Label Classification) for assigning medical codes to the clinical text. Generated word embeddings using CBOW word2vec methods over the preprocessed discharge summaries from MIMIC II and MIMIC III. In the proposed architecture, CNN gathers the information from the discharge summary and attention mechanism select relevant text fragments for the possible codes. Evaluation on the discharge summaries of MIMIC datasets showed the superior performance of the model. Further, the model also provides the satisfactory code assignments by making use of attention mechanism.

In general, the semantic meaning of a word depends on the words in the context. However, in biomedical domain, the semantic meaning of word is influenced by chunks and named entities in the context. Jiang et al. [37] proposed a simple and novel method based on skip gram model to jointly learn the word embeddings for words, chunks and named entities from Medline documents. Experimental results showed that the proposed embedding outperformed the general skipgram, cbow and glove embeddings.

Pyysalo et al.[39] created first set of embeddings in biomedical domain using the entire biomedical literature available (by September 2013) from PubMed and PubMed Central Open Access(PMC OA). Initially, collected abstracts from PubMed and full text from PMC OA, removed the duplicates forming a corpus of 5.5 billion tokens. Using word2vec, generated three different word embeddings from PubMed abstracts, PMC OA full texts and a combination of PubMed abstracts and PMC OA full texts. Evaluation on three biomedical domain corpora in the task of named entity recognition showed the effectiveness of generated embeddings.

Shweta et al.[40] studied Elman [98] and Jordan [99] RNN architectures to the task of patient data de-identification in clinical records. Viewed the problem as sequence labelling task where the proposed method predicts the appropriate labels for the words. Experimented with different word embeddings like RNNLM [86] trained on Broad News Corpus and CBOW trained on Google news corpus. Evaluation on 2014 i2b2 challenge data set [100] showed that both RNN architectures outperformed the CRF baseline. Further, CBOW embeddings gave better performance compared to RNNLM embeddings.

Zhao et al. [41] proposed a novel method which includes text skeleton and RNN for the patient data de-identification. Text skeleton represents the structure of the medical record. The proposed model consists of two branches with each having RNN layer and embedding layer. The outputs of both the branches are taken by the softmax layer to predict the labels. Here, the word embeddings were learned in an unsupervised way and then fine-tuned during the training. Experimental results showed that model performed well on three different standard datasets indicating the strong generalization ability of the model.

Dernoncourt et al. [49] proposed LSTM based architecture with character and word embeddings for patient data de identification. The character enhanced token embeddings layer maps token to feature vectors which is obtained by the concatenation of two embeddings. The first one directly encodes token to vector while second one comes from character level token encoder. The token embeddings are pre initialized with Glove embeddings trained on Wikipedia. The label prediction layer with the sequence of embeddings as input, outputs the sequence of probability vectors and finally sequence optimization layer outputs the sequence of labels. The proposed model with character and word embeddings evaluated on MIMIC III and i2b2 challenge datasets outperformed the baselines.

Nikolenko et al. [51] viewed the task of extracting adverse drug reactions as a sequence labelling problem and explored Bi-LSTM CRF model for that. Used the word embeddings generated from 2.5 million health related reviews[17]. With the health2vec word embeddings as input, Bi-LSTM encode the words into fixed length vectors and CRF operates on these vectors to assign the labels. Evaluation on CSIRO Adverse Drug Event Corpus (CADEC) corpus having reviews from AskaPatient.com[91], demonstrated the effectiveness of the model. Further, the use of character level embeddings generated using CNN [101] along with health2vec embeddings improved the performance.

Kirk Roberts [60] conducted experiments to assess corpus size vs similarity trade-off in the generation of clinical word embeddings. In general, the corpus size is one of the crucial factors influencing the quality of word embeddings. However, in clinical NLP, only small data sets are available because of restrictions to access EHR data and this forces a choice between small but representative corpora and large but unrepresentative corpora for inferring word embeddings. To assess the corpus size vs similarity trade off, conducted two experiments (i2b2 2010 concept and assertion tasks) using RNN and CNN model respectively. The experiments make use of word embeddings inferred independently using word2vec CBOW from six corpora ranging from small size representative i2b2 corpus to large size unrepresentative Google News Corpus. Empirical results show that combination of corpora are effective, but the single best corpus is generally task-dependent.

Finley et al. [61] studied embeddings induced from various corpora to study of effect of genre of corpus in representing medical terms. Using word2vec CBOW, generated separately from three corpora (Clinical Notes, PubMed Central Articles and Wikipedia). Evaluated the generated word embeddings using the tasks of semantic similarity and relatedness, clinical document retrieval and biomedical word sense disambiguation task. Empirical results showed that generating word embeddings from domain corpora results in improvement in performance compared to the word embeddings from a related, but not domain corpora.

Liu et al. [112] studied the use of LSTM for the recognition of clinical entities and protected health information. LSTM consists of three layers. First layer maps each word in the sequence to fixed length vector which is a concatenation of word level and character level embedding, the second layer output the vector sequence having the context information and finally the last layer assigns the labels to each word. Here the word level embedding and character level embedding are inferred using word2vec and CNN or Bi-LSTM. Results showed that use of character level embeddings along with word level embedding improved the performance.

Lee et al. [114] proposed LSTM based framework for patient data de-identification which makes use of human engineered features besides dense vector representations. For each token, a binary feature vector having all the human engineered features is computed. With this binary feature vector as input, feed forward neural network outputs a vector which is concatenated with word and character level embeddings of the token. Here the word embeddings are pre-trained Glove embeddings and character embeddings are generated by Bi-LSTM. Results showed that addition of human engineered features improved the performance of the model.

Xie et al. [115] proposed a deep learning model based on Bi-LSTM to identify e-cigarette components and adverse events in social media text. Generated the word embeddings using skipgram of word2vec over the corpus collected from E-cigarette forum. With the word embeddings as input, Bi-LSTM model identified the entities with a recall of 91.80% and outperformed all the baselines.

Chalapathy et al. [116] explored the use of Bi-LSTM CRF for clinical concept extraction with embeddings inferred from out of domain corpus. Experimented with pre-trained word2vec embeddings inferred from Google News Corpus and Glove embeddings inferred from general domain corpus. Evaluation on 2010 i2b2 V/A challenge showed that the model outperformed all the recent approaches and was close to the best submission from the original challenge. Use of embeddings inferred from domain specific corpus like MIMIC III can improve the performance further.

Chalapathy et al. [117] experimented with Elman and Jordan RNN architectures and Bi-LSTM CRF models for identifying drug names. All the models are initialized with embeddings randomly sampled with in the range [-1,1]. Experimental results showed that Bi-LSTM model outperform other models and ranked close to the start of art system in SemEval 2013 Task 9.1. Use of pre-trained word2vec or glove embeddings or embeddings inferred from domain specific corpus like MIMIC III can improve the performance further.

Unanue et al. [119] proposed Bi-LSTM CRF with specialized word embedding and hand crafter features as input, for identifying health related named entities. Specialized word embedding is obtained by the concatenation of pre-trained Glove embeddings inferred from common crawl, MIMIC III embeddings inferred using Glove and character embeddings level word embeddings generated using Bi-LSTM. Hand crafted features include morphological, semantic and word embedding clusters. Experimental results showed the effectiveness of specialized embeddings and also hand features doesn't improve the performance much.

Cocos et al. [120] experimented with Bi-LSTM model for identifying and classifying adverse drug reactions in tweets. Evaluated the model with i) randomly initialized embeddings ii) fixed or updated pre trained embeddings inferred from 400 million general tweets [28]. Results showed that Bi-LSTM model with fixed embeddings outperformed all others and further, the model achieved optimal performance with fewer training examples.

Henghui et al. [122] used Bi-LSTM CRF model with domain specific ELMo embeddings as input, for the extraction of clinical concepts like problems, treatments and tests. As the Pretrained ELMo embeddings were induced from out of domain corpus, induced domain specific ELMo embeddings from a corpus having medical related Wikipedia pages, discharge summaries and radiology reports from MIMIC III. Experimental results showed that domain specific ELMo embeddings boosted the performance of the model.

Yifeng et al. [126] experimented with LSTM CRF model for the task of clinical concept extraction. The hidden state of LSTM with pretrained ELMo embeddings as input is concatenated with low dimensional representation of ontology based features and then fed into CRF to get the label. Here, sparse and high dimensional features built from medical ontology are mapped to low dimensional vector using a dense layer with dropout. Evaluation showed that use of ELMo embeddings and ontology based features improved the performance of the model.

5.2 **Code Embeddings**

EHRs contain patient information in the form of free text as well as medical codes. Medical codes are used to ensure consistency in recording the patient information. Dense vector representations of the medical codes finds applications in various tasks in health care analytics.

Edward Choi et al. [8] proposed a GRU based framework for early detection of heart failure. Traditional machine learning methods use aggregation of medical concepts (diagnosis codes, procedure codes, medical codes) in an observation window as features without considering the sequence of events in the observation window. Used skip gram technique to learn the fixed length vectors representations of the medical events like diagnosis codes, procedure codes and medication codes and GRU to model the temporal relations among the medical events. Experimental results show the improved performance of GRU model compared to the machine learning techniques and demonstrate the importance of temporal information.

Youngduck Choi [9] applied skip gram algorithm on a private medical claims data set to embed all the medical concepts (diagnosis codes, procedure codes, laboratory codes, drug codes) in the same continuous vector space. High quality representations of medical concepts find applications in various tasks in health care analytics. For example, as all the medical concepts are embedded in the same space, a medication to a disease can be found by finding the nearest neighbors to the specific disease code.

Xiangrui Cai et al.[43] proposed an architecture based on CBOW model with attention to jointly learn the embedding and temporal scope for each medical concept. The proposed model uses the attention mechanism to learn the temporal scope for medical concepts. The model is evaluated on two datasets in the tasks of clustering and nearest neighbors search and the results obtained are competitive.

Zhengping Che et al. [47] studied CNN with medical event embeddings in the task of risk prediction. A corpus of medical events is obtained from a real EHR data set. Used CBOW Word2Vec with a context windows of size 20 to generate the representations of size 200 for all the events with atleast a frequency of five. In the prediction of diabetes and congestive failure, the proposed model outperformed the baselines SVM and Logistic Regression(LR).

## 5.3 CUI Embeddings

UMLS consists of Metathesaurus, Semantic Network, Specialist Lexicon and a set of software tools like Metamap. Metathesaurus consists of over 5 million concept names incorporated from 100 controlled vocabularies like SNOMED CT, ICD, Rx-Norm etc. It clusters all the synonym concepts from different vocabularies into a single concept and assigns a unique identifier called Concept Unique Identifier. CUI maps synonym concept names in different vocabularies and thus it acts as a mapping structure between controlled vocabularies. CUI consists of 8 characters starting with C followed by 7 digits. For example, CUI of head pain is C0018681. Mapping CUIs to dense vector representations and use of CUI embeddings includes the domain knowledge from the controlled vocabularies.

Choi et al.[9] learned representation of UMLS concepts using co-occurrence counts of concepts with in fixed time intervals derived from clinical narratives [102]. These outperformed the embeddings learned by De Vine et al.[13] in the tasks of medical relatedness and similarity.

De Vine [13] presented concept based Skip-gram model to map UMLS CUIs to dense vector representations. In contrast to the traditional skip-gram which learns embeddings for words, this model learns the embeddings for UMLS concepts extracted from clinical records and medical journal abstracts. Using MetaMap v11.2 [103], converted free text into sequence of concepts and then applied skip gram model. Evaluation on two medical word similarity datasets showed the effectiveness of UMLS CUI embeddings. This model finds applications in Information Retrieval and Analytics related to clinical domain.

## 5.4 Augmented Embeddings

Word embeddings are induced from unlabeled corpus and the size of corpus is one of the determining factors in the quality of the inferred embeddings. The size of corpus is limited in clinical domain. In order to make up for the size of corpus and improve the quality of inferred embeddings, domain specific as well as task specific information can be included. Some of the possible ways to include information is tuning pre trained off-the-shelf word embeddings [2] or make use of domain knowledge from ontologies like UMLS [5, 36, 44 and 53].

### 5.4.1 Augmented Word Embeddings

Patel et al.[2] adapted CBOW algorithm proposed by Mikolov et al. [62] to include task and domain specific information to the pre-trained word embeddings in order to improve their utility in medical coding, which is a domain specific task. Information added include the coding data from private medical claims data set and the information from ICD-10 ontology. Evaluated the approach on five different pre-trained word embeddings which include GNews pretrained embeddings and the remaining embeddings proposed by Pyysalo et al. [39]. Results demonstrated the better performance of modified word embeddings on a private medical claims dataset and proved that adding domain specific information is useful.

Boag et al. [5] explored the addition of domain knowledge to word embeddings to make up for the small sizes of the clinical corpus. Levy et al. [104] introduced Word2Vecf (an extension to Word2Vec which uses dependency based contexts instead of linear contexts.) Generated word embeddings with word2vecf using the corpus having (w,c) pairs and (w,CUI) pairs generated from MIMIC III and UMLS respectively. Here 'w' represents word, 'c' represents the context and CUI represent Concept Unique Identifier. Evaluation on MayoSRS dataset showed the superior performance of augmented word embedding compared to the baselines.

Ling et al. [36] explored the idea of improving word2vec model with the addition of domain knowledge in the form of a graph using graph regularization. Initially generated a weighted graph from knowledge base such that related word, opposite words, weakly related words are connected by edges with the weight values 1, -1 and 0.5 respectively. After that, integrated the graph into original word2vec using graph regularization. Evaluation on four general data sets and two biomedical NLP tasks showed better performance of the model.

Eneldo et al. [55] adapted All-in-Text proposed by Nam et al. [105] to produce word embeddings. The method makes use of both word-occurrence information as well as the association between the labeled documents and their labels from sources such as PubMed. Inspired by paragraph2vec model [89], All-in -Text jointly learns the embeddings for words, documents and the labels. The model is trained on BioASQ data set consisting of title and abstracts of PubMed articles labeled with associated Medical Subjet Headings (MeSH) Ontology terms. Evaluation on two similarity and relatedness datasets demonstrated the effectiveness of the model.

Bin He et al. [56] proposed CNN with multi pooling operation for medical relation classification. The introduced multi pooling operation is good at capturing the position information of local features relative to the concept pair which is crucial for relation classification. The model makes use of word representation which is obtained by the concatenation of three embeddings namely word embedding and the two positional word embeddings. Word embeddings were generated using word2vec [62] and MIMIC III [78] corpus. Evaluation on i2b2 2010 corpus [92] demonstrated substantial improvement over the baselines.

5.4.2 **Augmented Code Embeddings**

Choi et al. [44] proposed GRAM- Graph based attention model to learn the medical concept embedding using knowledge from medical Ontology and electronic medical records. The proposed model represents a medical concept as a weighted sum of its ancestors in ontology using attention mechanism so that representations learned by the model aligns with the medical knowledge. Glove [87] algorithm was used to get embeddings of the ancestors. The predictive model along with attention was trained in an end to end manner and evaluated on two sequence prediction tasks and one heart failure prediction task. Results demonstrate superior performance of the model.

Feng et al. [53] proposed MG-CNN, a deep learning based framework to study the influence of patient characteristics on total hospital costs and Length of stay (LOS). The framework uses multi-granularity embedding method which adds the prior medical knowledge from the four levels of ICD Ontology and maps the medical concepts to fixed length vectors. Initially, four fine grained codes are obtained for each ICD code in the data set and then sentences are constructed in four levels. Skip-gram operates on these sentences and generates the medical concept embeddings at each level. Experimental results demonstrate the superior performance of the model in prediction tasks for patient with rare medical events. Further, visualization using t-SNE [106] algorithms showed that related medical concepts are closer.

5.7 **Patient Embeddings**

Patient information right from admission to discharge is recorded in EHR using free text as well as medical codes. Dense vector representation of patients is required in applications like prediction of clinical events and next visit time [33], prediction of unplanned readmission [30] etc. Patient embeddings encode the patient information recorded in free text [52] or medical codes [7, 33, 48, 54, 58 ] or both [38] into dense fixed length vectors.

Choi et al.[9] achieved better predictive performance using the visit representations obtained by the aggregating code representations in the visit .However, it not optimal as it completely ignores sequential relationship among the visits. Edward Choi et al. [7] proposed a simple and scalable two layer neural network based architecture to learn dense and interpretable representations of both medical codes and patient visits using large real EHR data sets. The model learns better representations with the inclusion of code co-occurrence as well as the patient visit sequence information. Experimental results show that learned representations are efficient with improved performances in two clinical prediction problems compared to several baselines.

Nguyen et al.[30] proposed Deepr, a CNN based model for medical records. Time gap between the patient visits is random and to handle this, time gaps and transfers are represented as special words. With this representation, patient is represented as sequence of phrases separated by special words where each phrase represents a visit. Generated embeddings using word2vec over a real EHR data. CNN encodes the sequence of codes into feature vector, which is then used by the classifier to predict the future risk. Evaluation shows the effectiveness of the proposed model in unplanned readmission.

Edward Choi et al.[33] proposed Doctor AI: GRU based predictive model that learns patient representations from real EHR data and predicts clinical events and next visit time. Generated code embeddings using word2vec[62] over the real EHR data collected from a private hospital. GRU model takes patient encounters (sequence of diagnosis and medication codes) as input and the output of hidden layers is the patient representation which is given to a softmax layer to predict the codes and ReLU to predict the next visit time. Experimental results showed the superior performance of the model as well as the adaptability of the model across hospitals.

Dligach et al. [38] proposed a simple and novel neural network architecture to low dimensional, dense representations of a patient from clinical text. The model takes a set of CUIs extracted from patient notes and predicts the billing codes. Initially, CUIs are mapped to embeddings which are averaged and passed on to a hidden layer. The output of hidden layer, a 1000 dimensional vector is used by the output layer to predict the billing codes. Here, the 1000 dimensional vector produced by hidden layer represents the patient. CUI embeddings are generated using word2vec over a corpus of CUIs extracted from MIMIC III [78] clinical notes. Experimental results demonstrated the effectiveness of learned patient representations.

Stojanovic et al. [48] proposed disease+procedure2vec (dp2v), a novel method based on skip gram model[62] to embed disease and procedure codes into the same low dimensional space. Each patient record is represented as a sequence of disease and procedure codes. Treating each patient record as a sentence and codes as words, skipgram model learns the representations of the disease and procedure codes. Generated the patient vector by adding the vectors of disease and procedure codes in each record. With the patient feature vectors as input, the classifier predicts the length of stay, total incurred charges and mortality rates.

Madhumita et al. [52] studied Stacked Denoising Autoencoder as well as Paragraph2vec models [89] to encode patient information in clinical notes into task-independent, dense fixed length vectors. Initially build a corpus having only the clinical notes of adult patient (>=18) with only one hospital admission and transformed it into bag of words and bag of CUIs. Then, patient representations are generated using the two unsupervised models. Feed forward neural networks with dense patient representations as input are evaluated on multi independent tasks to demonstrate the task-independent nature of patient representations. Results showed that there is not much difference in performance while using bag of CUIs compared to bag of words.

Zhu et al. [54] proposed deep learning based framework to find the clinical similarities between patients in a way that EHR temporal properties are preserved. Initially, concept embeddings are generated using word2vec[62] over a real EHR dataset. Using the generated embeddings, medical events of patients are mapped to fixed length vectors. Finally, patient representation is obtained by stacking all the fixed length vectors of the medical events as a matrix. To obtain the patient similarity, studied both unsupervised methods like RV coefficient[107], dCov efficient [108] and supervised methods like CNN. Evaluation on real EHR dataset showed the superior performance of the proposed framework.

Dubois et al. [58] studied two methods embed-and-aggregate and RNN to learn patient representations that can be used in the prediction of various clinical events. In embed-and-aggregate method, aggregate operation on code embeddings results in note level representation and further aggregation of note level representation results in patient representations. Here aggregate operation is an element-wise min, max or mean. Code embeddings are obtained using Glove[87] and Stanford EHR data. In second method, RNN [98] is applied to learn the representations of concepts, notes and patients simultaneously from Stanford EHR data. Results showed that embed-and-aggregate embeddings is competitive with RNN embeddings and moreover, it is simple to compute.

Miotto et al. [59] proposed a novel unsupervised deep learning method to infer patient representations from a real EHR data which can improve medical events prediction. The proposed model is a multi-layer neural network comprising of denoising autoencoders. Each layer of denoising autoencoder produce a more abstract representation than the previous level by applying more non-linear operations and the output of the final layer results in the patient representations referred to as "Deep Patient". The generated representations are evaluated by predicting the probability of patients to develop various diseases. Results showed that the learned patient representations outperformed other methods.

5.8 **Phrase Embeddings**

Phrase embeddings maps phrases to fixed length dense vectors. Phrase embeddings can be generated from the aggregation of embeddings of words in the phrases or directly using word2vec or paragraph2vec models.

Limsopatham et al. [113] experimented with phrase based MT (Koehn et al. 2003) for the task of medical concept normalization. Initially, similarity score is computed between twitter phrase and the description medical concept and then cosine similarity is calculated between twitter phrase and the description. Finally, twitter phrase is mapped to medical concept based on the linear combination of two the similarity scores. Here they have experimented with word embeddings generated using word2vec, glove on drug related tweets dataset and BioMed Central literature dataset.

Henry et al. [121] studied various dimensionality reduction techniques like skipgram, cbow, svd, explicit co-occurrence vectors and various aggregation multi term aggregation methods like sum, average, direct construction using compoundify tool or Meta Map in the context of semantic relatedness. Evaluation showed that i)none of the multi term aggregation method is better than the other while gives the flexibility in choosing the methods ii) cbow embeddings with a size of 200 outperformed others.

5.9 **Sentence Embeddings**

Sentence embedding maps sentence to fixed length dense vectors. Sentence embedding can be generated from the aggregation of embeddings of words in the sentence or directly using paragraph2vec model.

Hughes et al. [24] explored convolutional neural networks with word embeddings for task of medical text classification at sentence level. Generated word embeddings from 15k research articles obtained from PubMed using word2vec[62]. Explored other embeddings like sentence embeddings, mean word embeddings and word embeddings with BOW features. Sentence word embeddings are generated using doc2vec[89]. Mean word embeddings (ZeroMean /EliminateMean) are calculated by taking the average of the word embeddings of all the words in the sentence. In ZeroMean, zero vectors are used for out of vocabulary words and in EliminateMean, words without embeddings are removed. Experimental results show that CNN with word embeddings outperformed all other methods.

Zhang et al. [50] proposed unsupervised deep learning based framework for identifying psychiatric symptoms in clinical text. Initially, the seed symptoms are collected various sources like MedLinePlus, Mayo Clinic and American Psychiatric Association. Generated sentence level and phrase level embeddings using paragraph2vec [89] from psychiatric notes from CEGS N-GRID 2016 Challenge [109], psychiatric forum data from WebMD and MIMIC II [77] incrementally. Using the generated embeddings, both seed symptoms and candidate symptoms are encoded as fixed length vectors. Finally, using the cosine based semantic similarity most probable psychiatric symptoms are identified. Evaluation on CEGS N-Grid data demonstrated the effectiveness of the model and revealed two interesting things. One is the performance of embeddings reduced with the use of MIMIC II data and sentence level embeddings outperformed the phrase level embeddings.

Luo et al. [57] proposed a deep learning framework to map disease and procedure names in Chinese Discharge Summaries to standard names. The proposed model makes use of tensor generator, multi-view CNN and multi-task framework. Tensor generator generates matching tensors for two mention-entity pairs for disease and procedure by making use of character, word and sentence embeddings. Character and word embeddings are generated using word2vec and Chinese medical corpora, sentence embeddings using Bi-LSTM. Multi view CNN operators on matching tensors and generate vectors based on which multi task framework produces the matching score.

5.10 **Document Embedding**

Document embedding maps documents to fixed length dense vectors. Document embedding can be generated from the aggregation of embeddings of sentences or directly using paragraph2vec model.

Wang et al. [25] explored the effectiveness of different embeddings like word2vec [62], glove [87], doc2vec [89] and the proposed Ariadne embeddings in different tasks. Ariadne is a count based word embedding generating method that applies random projection over the global co-occurrence matrix. Applied CBOW from word2vec, glove and doc2vec algorithms to the corpus having Medline articles to generate word and document embeddings. In word analogy task, word2vec embeddings outperformed the proposed word embeddings. In Information Retrieval task, the proposed word embeddings performed as wells as doc2vec embeddings. Results shows that one has to consider the requirements while choosing one of the many word embedding models.

Baumel et al.[42] proposed Hierarchical Attention –bidirectional Gated Recurrent Unit(HA-GRU) model for the task of assigning ICD codes to medical text. The proposed model consist of two GRU layers. The first GRU layer encodes the input sentences into fixed length vector and the second GRU layer encode all these sentence vectors into another fixed length vector which represents the documents. Finally with the document vector as input, the classifier which is a fully connected layer with softmax assigns the labels. Evaluation on the discharge summaries of MIMIC II [77] and MIMIC III [78] datasets shows the effectiveness of the proposed model. Results shows that tokenization and hierarchical segmentation contributed to the improved performance of the model.

Li et al. [46] proposed DeepLabeler, a deep learning based method to assign ICD-9 codes to medical discharge summaries. DeepLabeler consists of CNN as well as Doc2Vec[89]. CNN through convolution and pooling operations encodes the medical record into feature vectors which contains the local features. Doc2Vec which is trained over the entire corpus, encodes the medical documents into document vectors which contains global features. Both the vectors are concatenated and given to a classifier which assign the ICD 9 codes. The novelty of this mode lies in the combination of local and global features which are crucial in the ICD 9 code assignment.

Banerjee et al. [118] proposed intelligent word embedding which combines semantic dictionary mapping and neural embedding to map radiology reports to dense vectors. Initially, radiology report is condensed using a series of preprocessing steps and using semantic dictionary mapping, words in corpus are mapped to terms in domain specific dictionary which reduces the size of vocabulary and handles the OOV issue. After that using word2vec, word embeddings are generated. Finally, the average of word embeddings gives the dense vector representation of the radiology report.

6. **EVALUATION OF EMBEDDINGS**

The quality of embeddings can be assessed in two ways namely intrinsic and extrinsic. Intrinsic evaluation looks how well the induced embeddings are able to encode syntactic and semantic information. Some of the tasks used in intrinsic evaluation are Clustering [43], Nearest Neighbor Search (NNS) [43], Similarity and Relatedness [5, 9, 13, 36 and 55]. Similarity and Relatedness are the most commonly used tasks in intrinsic evaluation. For example, the datasets used in the word similarity or word relatedness evaluation tasks like UMNSRS Similarity [131], UMNSRS Relatedness [131], MayoSRS [132], Pedersen's dataset [133], Hliaoutakis's dataset [124] consists of word pairs along with a similarity or relatedness score assigned by medical experts. Correlation between the cosine similarity of word vectors and expert assigned score is a measure of the quality of embeddings [5, 36 and 55]. Table VIII shows the number of words pairs in each of the word similarity or relatedness datasets.

In extrinsic evaluation methods, embeddings are used as input features in downstream tasks like Named Entity Recognition, Medical Concept Normalization, Medical Text Classification etc. and improvement in the performance of the model is a measure of the quality of embeddings. Extrinsic evaluation is necessary to find the effectiveness of embeddings in real world tasks. Table IX shows the intrinsic and extrinsic evaluation tasks in various clinical embeddings

**TABLE VII**

**SUMMARY OF CLINCAL NLP TASKS**

| Task | Research Paper | Type of Embedding | Embedding Model | Corpus |
|---|---|---|---|---|
| Clinical Abbreviation Expansion | [3] | Word Embeddings | Word2Vec | Clinical Notes, Wikipedia Articles, ICU related Books and papers |
| Medical Concept Normalization | [4] | Word Embeddings | Word2Vec | Merriam-Webster Thesaurus, Merriam-Webster Medical Dictionary, Clinical Text, Health Related Tweets |
| | [10] | | | Health related Reviews, PubMed Literature |
| | [11] | | | Google News Corpus, BioMed Literature |
| | [12] | | | Google News Corpus, Generic Tweets and Drug Related Tweets |
| | [57] | Character, Word and Sentence Embeddings | Word2Vec and Bi-LSTM | Chinese Medical Corpora |
| Text Classification | [15] | Word Embeddings | Word2Vec | Two private EMR datasets and medical text book (the 7th edition of internal medicine) |
| | [16] | Word Embeddings and CUI Embeddings | | MIMIC III |
| | [19] | Word Embeddings | Word2Vec and Glove | Medical and Generic Corpus |
| | [22] | Word Embeddings | Word2Vec | Google News Corpus and Medical Corpus |
| | [24] | Word Embeddings and Sentence Embeddings | Word2Vec and Doc2Vec | PubMed Literature and Merck Manuals |
| | [46] | Word and Document Embeddings | | MIMIC II and MIMIC III |
| Named Entity Recognition | [17] | Word Embeddings | Word2Vec | Health related Reviews, PubMed Literature |
| | [18] | | | Drug related Reviews and Tweets |
| | [26] | | | PubMed articles, Wikipedia articles and clinical notes |
| | [27] | | | PubMed articles, Wikipedia articles and clinical notes |
| | [31] | | Word2Vec and FastText | Wikipedia, MIMIC III, PubMed and medical corpus |
| | [51] | | Word2Vec | Health Related Reviews |
| ICD Coding | [2] | Augmented Word Embeddings | Word2Vec | Google News Corpus, PubMed Literature and Wikipedia |
| | [20] | Word Embeddings | FastText | MIMIC III |
| | [21] | | Glove | MIMIC III |
| | [23] | | Word2Vec | Health Related Reviews, PubMed Literature, Wikipedia and Google News Corpus |
| | [28] | | Glove | Common Crawl |
| | [29] | | Word2Vec | Google News Corpus |
| | [35] | | | MIMIC II and MIMIC III |
| | [42] | | | MIMIC II and MIMIC III |
| | [46] | Word and Document Embeddings | Word2Vec and Doc2Vec | MIMIC II and MIMIC III |
| Information Retrieval | [25] | Word and Sentence Embeddings | Word2Vec, Glove and Doc2Vec | MEDLINE Articles |
| Clinical Predictions | [8] | Code Embeddings | Word2Vec | Private EHR dataset |
| | [30] | Patient Embeddings | | Private EHR dataset |
| | [33] | | | Private EHR dataset |
| | [47] | Code Embeddings | | Private EHR dataset |
| | [53] | Patient Embeddings | | Private EHR dataset |
| | [59] | Patient Embeddings | Stacked Denoising Auto Encoders | Private EHR dataset |
| Relation Classification | [32] | Word Embeddings | Word2vec | Google News Corpus and MIMIC III |
| | [56] | | | MIMIC III |
| | [40] | Word Embeddings | Word2Vec, RNNLM | Google News Corpus and Broadcast News Corpus |
| Patient De-identification | [41] | | RNN Encoder and Decoder[22] | Medical Corpus |
| | [49] | | Glove | Wikipedia |
| Patient Similarity | [54] | Patient Embeddings | Word2Vec | Private EHR Dataset |

TABLE VIII

WORD SIMILARITY /RELATEDNESS DATASETS

| Dataset | # Word Pairs | Word Similarity / Relatedness |
|---|---|---|
| UMNSRS Similarity [131] | 566 | Word Similarity |
| UMNSRS Relatedness [131] | 588 | Word Relatedness |
| MayoSRS [132] | 101 | Word Relatedness |
| Pedersen's dataset [133] | 30 | Word Similarity |
| Hliaoutakis's dataset [124] | 34 | Word Similarity |

TABLE IX

INTRINSIC AND EXTRINSIC EVALUATION IN CLINICAL EMBEDDINGS

| Type of Embedding | Intrinsic Evaluation | Extrinsic Evaluation |
|---|---|---|
| Character | - | MCN[111] |
| Word | Similarity and Relatedness [34] | Clinical Abbreviation Expansion [3]<br>Medical Concept Normalization [4, 10, 11 and 12]<br>Medical Text Classification [15, 19 and 22]<br>Named Entity Recognition [17,18,26, 27, 31 and 51]<br>ICD Coding [20, 21, 23, 28, 29, 35 and 42]<br>Relation Classification [32 and 56]<br>Patient De-identification [40, 41 and 49] |
| Code | Similarity and Relatedness [9] Clustering and Nearest Neighbor Search(NNS)[43] | Heart Failure Detection[8]<br>Risk Prediction[47] |
| CUI | Similarity and Relatedness[9]<br>Similarity[13] | - |
| Augmented WE | Relatedness[5]<br>Similarity and Relatedness[36 and 55] | Medical Coding[2]<br>Biomedical IR[36]<br>Medical Relation Classification[56] |
| Augmented CE | - | Sequence prediction task and Heart failure prediction task [44]<br>Prediction of total hospital costs and length of stay (LOS) [53] |
| Patient Embeddings | - | Clinical Predictions[9]<br>Unplanned Readmission[30]<br>Medical Event Prediction[59]<br>Prediction of clinical events and Next Visit time[33]<br>Comorbidity detection task [38]<br>Prediction of length of stay, total incurred charges and mortality rates [48] |
| Phrase Embeddings | Similarity and Relatedness [121] | Medical Concept Normalization [113] |
| Sentence Embeddings | - | Medical Text Classification[24]<br>Identifying psychiatric symptoms[50]<br>Medical Concept [57] |
| Document Embeddings | - | ICD Coding [42 and 46]<br>Radiology Report Classification[118] |

## 7. DISCUSSION

In this section, we discuss various aspects related to embeddings and highlight the possible solutions from the surveyed research papers to various challenges in embeddings.

### 7.1 Genre of Corpus

The semantic information captured by word embeddings depends on the genre of the corpus from which they are induced. The use of language and the expertise of the author varies from corpus to corpus. For example, health discussion forums contains medical terms in colloquial language authored by general public whereas PubMed contains medical terms in professional language authored by researchers. So, embeddings induced from health forum can better model the colloquial medical terms where as embeddings induced from PubMed can better model the professional medical terms [1].

### 7.2 Size of Corpus

Embeddings are induced from large unlabeled corpus and the size of corpus is one of the factors which influence the quality of the inferred embeddings. A large corpus results in better coverage of vocabulary while a domain related corpus results in better representation of terms by capturing more semantic information. In clinical domain, the available corpora are small in size compared to general corpora. So, there is a tradeoff between large unrepresentative corpus and small representative corpus. Possible solutions to make up for the small of clinical corpora are defined below

#### 7.2.1 Use of combined corpora

Medical domain related text is available in various sources like Wikipedia, Medical Thesaurus, Medical Dictionary, PubMed and PMCOA, Clinical Notes from EHR. As the availability of large amount of text from a single source is not there, combining text from various sources and inferred embeddings from the combined corpora can potentially improve the quality of the embeddings induced [4, 26, 27 and 50].

#### 7.2.2 Inclusion of Domain Knowledge

Domain knowledge can be added either during inferring or updating embeddings as described below

##### 7.2.2.1 Inclusion of Domain knowledge while inferring embeddings

UMLS Metathesaurus clusters the concepts from over 100 controlled vocabularies like SNOMED CT, Rx-Norm etc and assigns all the concepts with same meaning, a CUI (Concept Unique Identifier). ICD contains disease and symptoms information, CPT contains procedures information, LOINIC contains laboratory observations information and Rx-Norm contains drugs related information. So, use of the information from these medical ontologies which contain abundant domain knowledge can make up for the size of corpora and improve the quality of the embeddings inferred [5, 13, 16, 36, 44, 53 and 55].

7.2.2.2 **Inclusion of Domain knowledge while updating embeddings**

Pre-trained embeddings released with embeddings models like Word2vec, Glove, FastText are trained on large unrepresentative corpora and as a result the quality of inferred embeddings is poor. However, as these embeddings are trained on large unlabeled corpora, they provide large coverage of vocabulary which we don't want to miss. So, pre-trained embeddings can be tuned in a supervised domain related task [2, 11, 29, 31 and 41] which incorporates the domain knowledge and improve the quality of embeddings.

7.2.3 **Ensemble of Word Embeddings**

Ensemble of word embeddings allow the model to make use of different information encoded in embeddings induced from different corpora. Pre-trained word embeddings provide large coverage of vocabulary while domain specific word embeddings better represent the terms. Further, the genre of medical corpora effects the word embeddings induced. For example, health discussion forum embeddings better model the colloquial medical terms while PubMed embeddings better models professional medical terms. So, using an ensemble of generic and domain specific word embeddings [12, 22] or an ensemble of embeddings induced from different genre of medical corpora improves the quality of embeddings [60].

7.2.4 **Semantic Features based on Word Embedding Clusters**

Word embeddings induced from small domain corpus can be improved further with the addition of clusters center vectors to the embedding of each word [15, 18]. Here adding cluster center vectors increase the amount of semantic information encoded in the vector representations and these vectors are obtained by applying any of the clustering algorithms on word embeddings.

7.3 **Multi Sense Embeddings**

A medical concept may have more than one meaning, referred to as multi sense. For example, aspirin can be used a medicine for both fever and cardiovascular disease. However models like word2vec, glove, fasttext assign a single representation ignoring the multi sense nature which reduces the quality of inferred embeddings and affects the performance of the model in downstream tasks. [7]- [9] generated embeddings for medical codes without considering the multi sense nature of the codes i.e., assigned only one representation for the codes with multi sense. [45] Using the context and topic information, learned the multi sense embeddings for the codes.

### 7.4 Combined Vector Space

Medical concepts namely diagnosis codes, procedure codes, laboratory codes and drug codes can be embedded into separate spaces. However, these concepts have a relation among them. For example, aspirin with the drug code '1191' is used to cure fever with the diagnosis code 'R50.80'. Having a separate embedding space results in difficulty in finding the relation between various medical codes. Advantage of having a combined vector space of all the medical codes is that, a medication to a disease can be found by finding the nearest neighbors to the specific disease code [9]. Further, learning the vector representations of medical words and codes jointly, improve the quality of both the vector representations [6]

### 7.5 Out of Vocabulary Words

It is common to have an induced vector space to have missing for words. For example, embeddings for domain specific words is missing in out of domain pre-trained embeddings (eg. phenacetin is not present in pre-trained vectors released by Google [2]). Further embeddings will be missing for rare words, misspelled words and different noun and verb forms even in the vector space inferred from domain related corpora. However, assigning vectors of randomly sampled values [11, 22 and 23] or zeros [12, 15 and 28] for such words is not an optimal solution because assigning such a representation ignores the meaning of the word. Possible solution for this is use of character embeddings [111] or contextualized embeddings [122] or use of semantic dictionary mapping [118].

### 7.6 Embedding Size

There is no rule of thumb to choose the optimal size of embeddings. Small embeddings are unable to capture much of the syntactic and semantic information, while large embeddings increase the complexity of the model. Most of the research studies tried embeddings of different sizes and then chose the dimension which gave the best results. However, generating and experimenting with embeddings of different sizes consumes lot of time and resources. So, there is a need of methods which find the optimal size of embeddings.

### 7.7 Ensemble of character and word embeddings

Character embeddings capture the morphological information like prefix, suffix and root besides having the embeddings for misspelled and rare words. However character embeddings doesn't capture the semantic information. Word embeddings treat word as an atomic unit due to which morphological information is lot and encode only the semantic information with no embeddings for misspelled words and rare words. So using an ensemble of character and word embeddings provides better vector representation [49, 51, 57, 112, 114 and 119].

### 7.8 Domain specific vs Off-the-shelf pretrained embeddings in Clinical NLP

Embeddings inferred from medical related corpus need not outperform off-the-shelf pretrained embeddings inferred from general corpus in any of the Clinical NLP tasks [11, 19 and 34]. For example, Google News embeddings outperformed BMC embeddings in the task of medical concept normalization [11]. The possible reason for this is that the corpus from which the pretrained embeddings are inferred includes domain related text also in the form of news articles and blog posts.

7.9 **Use of Off-the-shelf pretrained embeddings in Clinical NLP**

Despite being inferred from out of domain corpus, off-the-shelf pretrained embeddings released with models like word2vec, glove can be used in clinical NLP tasks owing to their ability to provide large coverage of vocabulary. These embeddings can be used a) by concatenating with domain specific embeddings [22,119] b) by adapting to clinical domain using the methods like fine tuning, linear mapping or nonlinear mapping [2,31] c) by using them as it is [11,19,28, 29, 40,49,116 and 120] without use of domain adaptation methods or concatenation with domain specific embeddings.

7.10 **Use of task specific information to improve embeddings**

The quality of domain specific embeddings can be improved further with the addition of task specific information which boots the performance of downstream task [2 and 3]. For example, in the task of medical coding [2], embeddings are improved with the addition of information from ICD-10 codes and in the task of clinical abbreviation expansion [3], the size of corpus is increased with the addition of Wikipedia articles of all candidate expansions of the abbreviations identified.

8. **CONCLUSION AND FUTURE WORK**

In this survey paper, we presented a detailed review of embeddings in Clinical Natural Language Processing. We classified medical corpora into four types, discussed each medical corpus in detail and then provided a comparison of them. We classified embeddings into nine types, discussed each embedding type in detail by highlighting the key aspects. We discussed various evaluation methods of embeddings which includes intrinsic as well as extrinsic methods. We discussed various aspects related to embeddings and highlighted possible solutions to the various challenges in embeddings. The possible solutions for the small size of corpus in clinical domain are use of combined corpora for inferring embeddings, inclusion of domain knowledge, ensemble of embeddings and generating semantic features from clusters of embeddings. The problem of missing embeddings for OOV words, misspelled words and different noun and verb forms can be solved using character embeddings or Elmo embeddings. Character embeddings lack semantic information, while word embeddings lack morphological information and hence using a combination of character and word embeddings can provide better vector representations.

Interpretability of embeddings, domain adaptation methods and methods to find the optimal size for embeddings are some of the promising research directions. Though embeddings capture prior knowledge in the form of syntactic and semantic information, it is still not known what each dimension represents. Poor interpretability of embeddings makes it a black box and limits its utility in domains where interpretability is a concern. Even though out-of-the shelf pretrained embeddings provide large coverage of vocabulary, there will be either no or poor quality embeddings for domain specific words. Not all the dimensions in an embeddings carry useful information and further use of large size embeddings require more computational resources and time.

**References**

[1] J. Huang, K. Xu, and V. V. Vydiswaran, "Analyzing multiple medical corpora using word embedding," in proceedings of IEEE International Conference on Health Informatics (ICHI), pp. 527–533, 2016.

[2] K. Patel, D. Patel, M. Golakiya, P. Bhattacharyya, and N. Birari, "Adapting pre-trained word embeddings for use in medical coding," in proceedings of BioNLP, pp. 302–306, 2017.

[3] Y. Liu, T. Ge, K. Mathews, H. Ji, and D. McGuinness, "Exploiting taskoriented resources to learn word embeddings for clinical abbreviation expansion," in proceedings of BioNLP, pp. 92–97, 2015.

[4] K. Lee, S. A. Hasan, O. Farri, A. Choudhary, and A. Agrawal, "Medical concept normalization for online user-generated texts," in proceedings of IEEE International Conference on Healthcare Informatics (ICHI), pp. 462–469, 2017,

[5] W. Boag and H. Kan´e, "Awe-cm vectors: Augmenting word embeddings with a clinical metathesaurus," CoRR, vol. abs/1712.01460, 2017.

[6] T. Bai, A. K. Chanda, B. L. Egleston, and S. Vucetic, "Joint learning of representations of medical concepts and words from ehr data," in proceedings of IEEE International Conference on Bioinformatics and Biomedicine (BIBM), pp. 764–769, 2017.

[7] E. Choi, M. T. Bahadori, E. Searles, C. Coffey, M. Thompson, J. Bost, J. Tejedor-Sojo, and J. Sun, "Multi-layer representation learning for medical concepts," in Proceedings of the 22Nd ACM SIGKDD International Conference on Knowledge Discovery and Data Mining, ser. KDD '16. ACM, pp. 1495–1504, 2016.

[8] E. Choi, J. Sun, A. Schuetz, and W. F. Stewart, "Using recurrent neural network models for early detection of heart failure onset," Journal of the American Medical Informatics Association, vol. 24, no. 2, pp. 361–370, 2016.

[9] Y. Choi, C. Y.-I. Chiu, and D. Sontag, "Learning low-dimensional representations of medical concepts," AMIA Summits on Translational Science Proceedings, p. 41, 2016.

[10] E. Tutubalina, Z. Miftahutdinov, S. Nikolenko, and V. Malykh, "Medical concept normalization in social media posts with recurrent neural networks," Journal of Biomedical Informatics, vol. 84, pp. 93 – 102, 2018.

[11] N. Limsopatham and N. Collier, "Normalising medical concepts in social media texts by learning semantic representation," in Proceedings of the 54th Annual Meeting of the Association for Computational Linguistics, pp. 1014–1023, 2016.

[12] M. Belousov, W. G. Dixon, and G. Nenadic, "Using an ensemble of linear and deep learning models in the smm4h 2017 medical concept normalisation task." in SMM4H@ AMIA, pp. 54–58, 2017.

[13] L. De Vine, G. Zuccon, B. Koopman, L. Sitbon, and P. Bruza, "Medical semantic similarity with a neural language model," in Proceedings of the 23rd ACM International Conference on Conference on Information and Knowledge Management, ser. CIKM, 2014, pp. 1819–1822.

[14] J. A. Minarro-Gimenez, O. Marin-Alonso, and M. Samwald, "Exploring the application of deep learning techniques on medical text corpora," Studies in health technology and informatics, vol. 205, pp. 584–588, 2014.

[15] Y. Shen, Q. Zhang, J. Zhang, J. Huang, Y. Lu, and K. Lei, "Improving medical short text classification with semantic expansion using wordcluster embedding," in Information Science and Applications. Springer, pp. 401–411, 2018.

[16] L. Yao, C. Mao, and Y. Luo, "Clinical text classification with rulebased features and knowledge-guided convolutional neural networks," in Proceedings of IEEE International Conference on Healthcare Informatics Workshop (ICHI-W), pp. 70–71, 2018.

[17] Z. Miftahutdinov, E. Tutubalina, and A. Tropsha, "Identifying diseaserelated expressions in reviews using conditional random fields," in Proceedings of International Conference on Computational Linguistics and Intellectual Technologies Dialog, vol. 1, no. 16, pp. 155–166, 2017.

[18] A. Nikfarjam, A. Sarker, K. O'Connor, R. Ginn, and G. Gonzalez, "Pharmacovigilance from social media: mining adverse drug reaction mentions using sequence labeling with word embedding cluster features," Journal of the American Medical Informatics Association, vol. 22, no. 3, pp. 671–681, 2015.

[19] N. Pattisapu, M. Gupta, P. Kumaraguru, and V. Varma, "Medical persona classification in social media," in Proceedings of the 2017 IEEE/ACM International Conference on Advances in Social Networks Analysis and Mining 2017, ser. ASONAM '17. ACM, pp. 377–384, 2017.

[20] C. Y. Li, D. Konomis, G. Neubig, P. Xie, C. Cheng, and E. P. Xing, "Convolutional neural networks for medical diagnosis from admission notes," CoRR, vol. abs/1712.02768, 2017.

[21] A. Karmakar, "Classifying medical notes into standard disease codes using machine learning," CoRR, vol. abs/1802.00382, 2018.

[22] N. Limsopatham and N. Collier, "Modelling the combination of generic and target domain embeddings in a convolutional neural network for sentence classification," in Proceedings of the 15th Workshop on Biomedical Natural Language Processing. Association for Computational Linguistics, pp. 136–140, 2016.

[23] Z. Miftahutdinov and E. Tutubalina, "Kfu at clef ehealth 2017 task 1: Icd-10 coding of english death certificates with recurrent neural networks."

[24] M. Hughes, I. Li, S. Kotoulas, and T. Suzumura, "Medical text classification using convolutional neural networks," Studies in health technology and informatics, vol. 235, pp. 246–250, 2017. [25] S. Wang and R. Koopman, "Semantic embedding for information retrieval," in BIR@ECIR, pp. 122–132, 2017.

[26] A. Jagannatha and h. yu, "Structured prediction models for rnn based sequence labeling in clinical text," in Proceedings of the 2016 Conference on Empirical Methods in Natural Language Processing. Association for Computational Linguistics, pp. 856–865, 2016.

[27] A. N. Jagannatha and H. Yu, "Bidirectional rnn for medical event detection in electronic health records," in Proceedings of the 2016 Conference of the North American Chapter of the Association for Computational Linguistics: Human Language Technologies. Association for Computational Linguistics, pp. 473–482, 2016.

[28] S. Ayyar and O. Bear, "Tagging patient notes with icd-9 codes," in Proceedings of the 29th Conference on Neural Information Processing Systems, 2017.

[29] X. Zhang, R. Henao, Z. Gan, Y. Li, and L. Carin, "Multi-label learning from medical plain text with convolutional residual models," arXiv preprint arXiv:1801.05062, 2018.

[30] P. Nguyen, T. Tran, N. Wickramasinghe, and S. Venkatesh, "Deepr: A convolutional net for medical records," IEEE Journal of Biomedical and Health Informatics, vol. 21, no. 1, pp. 22–30, Jan 2017.

[31] D. Newman-Griffis and A. Zirikly, "Embedding transfer for low-resource medical named entity recognition: A case study on patient mobility," in Proceedings of the BioNLP'18 workshop. Association for Computational Linguistics, pp. 1–11, 2018.

[32] Y. Luo, "Recurrent neural networks for classifying relations in clinical notes," J. of Biomedical Informatics, vol. 72, pp. 85–95, 2017.

[33] E. Choi, M. T. Bahadori, A. Schuetz, W. F. Stewart, and J. Sun, "Doctor ai: Predicting clinical events via recurrent neural networks," in Proceedings of the 1st Machine Learning for Healthcare Conference, vol. 56. PMLR, pp. 301–318, 2016.

[34] Y. Wang, S. Liu, N. Afzal, M. Rastegar-Mojarad, L. Wang, F. Shen, P. Kingsbury, and H. Liu, "A comparison of word embeddings for the biomedical natural language processing," Journal of Biomedical Informatics, vol. 87, pp. 12 – 20, 2018.

[35] J. Mullenbach, S. Wiegreffe, J. Duke, J. Sun, and J. Eisenstein, "Explainable prediction of medical codes from clinical text," in Proceedings of the Conference of the North American Chapter of the Association for Computational Linguistics: Human Language Technologies. Association for Computational Linguistics,
pp. 1101–1111, 2018.

[36] Y. Ling, Y. An, M. Liu, S. A. Hasan, Y. Fan, and X. Hu, "Integrating extra knowledge into word embedding models for biomedical nlp tasks," in Proceedings of International Joint Conference on Neural Networks (IJCNN), pp. 968–975, 2017.

[37] Z. Jiang, L. Li, D. Huang, and L. Jin, "Training word embeddings for deep learning in biomedical text mining tasks," in Proceedings of IEEE International Conference on Bioinformatics and Biomedicine (BIBM), pp. 625–628, 2015.

[38] D. Dligach and T. Miller, "Learning patient representations from text," in Proceedings of the Seventh Joint Conference on Lexical and Computational Semantics. Association for Computational Linguistics,
pp. 119–123, 2018.

[39] S. Moen and T. S. S. Ananiadou, "Distributional semantics resources for biomedical text processing," in Proceedings of the 5th International Symposium on Languages in Biology and Medicine, Tokyo, Japan, pp. 39–43. 2013.

[40] S. Yadav, A. Ekbal, S. Saha, and P. Bhattacharyya, "Deep learning architecture for patient data de-identification in clinical records," in Proceedings of the Clinical Natural Language Processing Workshop (ClinicalNLP). The COLING 2016 Organizing Committee, pp. 32–41, 2016.

[41] Y.-S. Zhao, K.-L. Zhang, H.-C. Ma, and K. Li, "Leveraging text skeleton for de-identification of electronic medical records," BMC Medical Informatics and Decision Making, vol. 18, no. 1, p. 18, 2018.

[42] T. Baumel, J. Nassour-Kassis, R. Cohen, M. Elhadad, and N. Elhadad, "Multi-label classification of patient notes: case study on icd code assignment," in Workshops at the Thirty-Second AAAI Conference on Artificial Intelligence, 2018.

[43] X. Cai, J. Gao, K. Y. Ngiam, B. C. Ooi, Y. Zhang, and X. Yuan, "Medical concept embedding with time-aware attention," Proceedings of the Twenty-Seventh International Joint Conference on Artificial Intelligence, 2018.

[44] E. Choi, M. T. Bahadori, L. Song, W. F. Stewart, and J. Sun, "Gram: Graph-based attention model for healthcare representation learning," in Proceedings of the 23rd ACM SIGKDD International Conference on Knowledge Discovery and Data Mining, ser. KDD '17. New York, NY, USA: ACM, pp. 787–795, 2017.

[45] F. Qian, C. Gong, L. Liu, L. Sha, and M. Zhang, "Topic medical concept embedding: Multi-sense representation learning for medical concept," in Proceedings of IEEE International Conference on Bioinformatics and Biomedicine (BIBM), pp. 404–409, 2017.

[46] M. Li, Z. Fei, M. Zeng, F. Wu, Y. Li, Y. Pan, and J. Wang, "Automated icd-9 coding via a deep learning approach," IEEE/ACM Transactions on Computational Biology and Bioinformatics, pp. 1–1, 2018.

[47] Z. Che, Y. Cheng, Z. Sun, and Y. Liu, "Exploiting convolutional neural network for risk prediction with medical feature embedding," arXiv preprint arXiv:1701.07474, 2017.

[48] J. Stojanovic, D. Gligorijevic, V. Radosavljevic, N. Djuric, M. Grbovic, and Z. Obradovic, "Modeling healthcare quality via compact representations of electronic health records," IEEE/ACM Transactions on Computational Biology and Bioinformatics, vol. 14, no. 3, pp. 545–554, 2017.

[49] F. Dernoncourt, J. Y. Lee, O. Uzuner, and P. Szolovits, "De-identification of patient notes with recurrent neural networks," Journal of the American Medical Informatics Association, 2016.

[50] Y. Zhang, O. Zhang, Y. Wu, H.-J. Lee, J. Xu, H. Xu, and K. Roberts, "Psychiatric symptom recognition without labeled data using distributional representations of phrases and on-line knowledge," J. of Biomedical Informatics, vol. 75, pp. 129–137, 2017.

[51] E. Tutubalina and S. Nikolenko, "Combination of deep recurrent neural networks and conditional random fields for extracting adverse drug reactions from user reviews," Journal of Healthcare Engineering, 2017.

[52] M. Sushil, S. Šuster, K. Luyckx, and W. Daelemans, "Patient representation learning and interpretable evaluation using clinical notes," Journal of Biomedical Informatics, vol. 84, pp. 103 – 113, 2018.

[53] Y. Feng, X. Min, N. Chen, H. Chen, X. Xie, H. Wang, and T. Chen, "Patient outcome prediction via convolutional neural networks based on multi-granularity medical concept embedding," in Proceedings of IEEE International Conference on Bioinformatics and Biomedicine (BIBM), pp. 770–777, 2017,.

[54] Z. Zhu, C. Yin, B. Qian, Y. Cheng, J. Wei, and F. Wang, "Measuring patient similarities via a deep architecture with medical concept embedding," in Proceedings of IEEE 16th International Conference on Data Mining (ICDM), pp. 749–758, 2016.

[55] E. Mencía, G. De Melo, and J. Nam, "Medical concept embeddings via labeled background corpora," in Proceedings of the 10th International Conference on Language Resources and Evaluation, LREC 2016. European Language Resources Association (ELRA), pp. 4629–4636, 2016.

[56] B. He, Y. Guan, and R. Dai, "Classifying medical relations in clinical text via convolutional neural networks," Artificial Intelligence in Medicine, 2018.

[57] Y. Luo, G. Song, P. Li, and Z. Qi, "Multi-task medical concept normalization using multi-view convolutional neural network," in AAAI, 2018.

[58] S. Dubois, N. Romano, D. C. Kale, N. Shah, and K. Jung, "Learning effective representations from clinical notes," arXiv preprint arXiv:1705.07025, 2017.

[59] R. Miotto, L. Li, B. A. Kidd, and J. T. Dudley, "Deep patient: an unsupervised representation to predict the future of patients from the electronic health records," Scientific reports, vol. 6, p. 26094, 2016.

[60] K. Roberts, "Assessing the corpus size vs. similarity trade-off for word embeddings in clinical nlp," in Proceedings of the Clinical Natural Language Processing Workshop (ClinicalNLP). The COLING 2016 Organizing Committee, pp. 54–63, 2016.

[61] S. V. S. Pakhomov, G. P. Finley, R. McEwan, Y. Wang, and G. B. Melton, "Corpus domain effects on distributional semantic modeling of medical terms," Bioinformatics, vol. 32 23, pp. 3635–3644, 2016.

[62] T. Mikolov, K. Chen, G. Corrado, and J. Dean, "Efficient estimation of word representations in vector space," in Proceedings of the International Conference on Learning Representations (ICLR 2013), pp. 1–12, 2013.

[63] F. Godin, B. Vandersmissen, W. De Neve, and R. Van de Walle, "Multimedia lab @ acl wnut ner shared task: Named entity recognition for twitter microposts using distributed word representations," in Proceedings of the Workshop on Noisy User-generated Text, pp. 146–153, 2015.

[64] A. Nikfarjam, A. Sarker, K. O'connor, R. Ginn, and G. Gonzalez, "Pharmacovigilance from social media: mining adverse drug reaction mentions using sequence labeling with word embedding cluster features," Journal of the American Medical Informatics Association, vol. 22, no. 3, pp. 671–681, 2015.

[65] H. Moen, F. Ginter, E. Marsi, L.-M. Peltonen, T. Salakoski, and S. Salanter¨a, "Care episode retrieval: distributional semantic models for information retrieval in the clinical domain," BMC Medical Informatics and Decision Making, vol. 15, no. 2, 2015.

[66] Y. Wu, J. Xu, M. Jiang, Y. Zhang, and H. Xu, "A study of neural word embeddings for named entity recognition in clinical text," in AMIA Annual Symposium Proceedings. American Medical Informatics Association, p. 1326, 2015.

[67] M. M. Ghassemi, R. G. Mark, and S. Nemati, "A visualization of evolving clinical sentiment using vector representations of clinical notes," in Proceedings of Computing in Cardiology Conference (CinC),

pp. 629–632, 2015.

[68] A. Henriksson, "Representing clinical notes for adverse drug event detection," in Proceedings of the Sixth International Workshop on Health Text Mining and Information Analysis. Association for Computational Linguistics, pp. 152–158, 2015.

[69] E. Choi, A. Schuetz, W. F. Stewart, and J. Sun, "Medical concept representation learning from electronic health records and its application on heart failure prediction," arXiv preprint arXiv:1602.03686, 2016.

[70] B. Shickel, P. J. Tighe, A. Bihorac, and P. Rashidi, "Deep ehr: A survey of recent advances in deep learning techniques for electronic health record (ehr) analysis," IEEE Journal of Biomedical and Health Informatics, vol. 22, no. 5, pp. 1589–1604, 2018.

[71] J. R. Firth, "A synopsis of linguistic theory, 1930-1955," Studies in linguistic analysis, 1957.

[72] Z. S. Harris, "Distributional structure," Word, vol. 10, no. 2-3, pp. 146– 162, 1954.

[73] D. Charles, M. Gabriel, and M. F. Furukawa, "Adoption of electronic health record systems among us non-federal acute care hospitals: 2008- 2012," ONC data brief, vol. 9, pp. 1–9, 2013.

[74] G. S. Birkhead, M. Klompas, and N. R. Shah, "Uses of electronic health records for public health surveillance to advance public health," Annual review of public health, vol. 36, pp. 345–359, 2015.

[75] T. Botsis, G. Hartvigsen, F. Chen, and C. Weng, "Secondary use of ehr: data quality issues and informatics opportunities," Summit on Translational Bioinformatics, vol. 2010, p. 1, 2010.

[76] P. B. Jensen, L. J. Jensen, and S. Brunak, "Mining electronic health records: towards better research applications and clinical care," Nature Reviews Genetics, vol. 13, no. 6, p. 395, 2012.

[77] M. Saeed, M. Villarroel, A. T. Reisner, G. Clifford, L.-W. Lehman, G. Moody, T. Heldt, T. H. Kyaw, B. Moody, and R. G. Mark, "Multiparameter intelligent monitoring in intensive care ii (mimic-ii): a publicaccess intensive care unit database," Critical care medicine, vol. 39, no. 5, p. 952, 2011.

[78] A. E. Johnson, T. J. Pollard, L. Shen, H. L. Li-wei, M. Feng, M. Ghassemi, B. Moody, P. Szolovits, L. A. Celi, and R. G. Mark, "Mimic-iii, a freely accessible critical care database," Scientific data, vol. 3, p. 160035, 2016.

[79] D. E. Rumelhart, G. E. Hinton, and R. J. Williams, "Learning representations by back-propagating errors," nature, vol. 323, no. 6088, p. 533, 1986.

[80] J. L. Elman, "Distributed representations, simple recurrent networks, and grammatical structure," Machine learning, vol. 7, no. 2-3, pp. 195–225, 1991.

[81] A. M. Glenberg and D. A. Robertson, "Symbol grounding and meaning: A comparison of high-dimensional and embodied theories of meaning," Journal of memory and language, vol. 43, no. 3, pp. 379–401, 2000.

[82] Y. Bengio, R. Ducharme, P. Vincent, and C. Jauvin, "A neural probabilistic language model," Journal of machine learning research, vol. 3, pp. 1137–1155, 2003.

[83] D. M. Blei, A. Y. Ng, and M. I. Jordan, "Latent dirichlet allocation," Journal of machine Learning research, vol. 3, pp. 993–1022, 2003.

[84] S. T. Dumais, "Latent semantic analysis," Annual review of information science and technology, vol. 38, no. 1, pp. 188–230, 2004.

[85] R. Collobert and J. Weston, "A unified architecture for natural language processing: Deep neural networks with multitask learning," in Proceedings of the 25th international conference on Machine learning. ACM, pp. 160–167, 2008.

[86] T. Mikolov, M. Karafiát, L. Burget, J. Cernocky̌, and S. Khudanpur, "Recurrent neural network based language model," in Eleventh Annual Conference of the International Speech Communication Association, vol. 2, p. 3, 2010.

[87] J. Pennington, R. Socher, and C. Manning, "Glove: Global vectors for word representation," in Proceedings of the conference on empirical methods in natural language processing (EMNLP), pp. 1532–1543, 2014.

[88] P. Bojanowski, E. Grave, A. Joulin, and T. Mikolov, "Enriching word vectors with subword information," Transactions of the Association for Computational Linguistics, vol. 5, pp. 135–146, 2017.

[89] Q. Le and T. Mikolov, "Distributed representations of sentences and documents," in International Conference on Machine Learning, pp. 1188–1196 ,2014.

[90] E. Grave, T. Mikolov, A. Joulin, and P. Bojanowski, "Bag of tricks for efficient text classification," in Proceedings of the 15th Conference of the European Chapter of the Association for Computational Linguistics, EACL, pp. 3–7, 2017.

[91] S. Karimi, A. Metke-Jimenez, M. Kemp, and C. Wang, "Cadec: A corpus of adverse drug event annotations," Journal of biomedical informatics, vol. 55, pp. 73–81, 2015.

[92] O. Uzuner, I. Goldstein, Y. Luo, and I. Kohane, "Identifying patient smok- ¨ ing status from medical discharge records," Journal of the American Medical Informatics Association, vol. 15, no. 1, pp. 14–24, 2008.

[93] N. Limsopatham and N. Collier, "Adapting phrase-based machine translation to normalise medical terms in social media messages," in Proceedings of the 2015 Conference on Empirical Methods in Natural Language Processing. Association for Computational Linguistics, pp. 1675–1680, 2015.

[94] R. Ginn, P. Pimpalkhute, A. Nikfarjam, A. Patki, K. O'Connor, A. Sarker, K. Smith, and G. Gonzalez, "Mining twitter for adverse drug reaction mentions: a corpus and classification benchmark," in Proceedings of the fourth workshop on building and evaluating resources for health and biomedical text processing. Citeseer, 2014.

[95] A. N´ev´eol, A. Robert, R. Anderson, B. Cohen, C. Grouin, T. Lavergne, C. Rondet, and P. Zweigenbaum, "CLEF eHealth 2017 Multilingual Information Extraction task Overview: ICD10 Coding of Death Certificates in English and French." in Workshop of the Cross-Language Evaluation Forum. Dublin, Ireland: CEUR-WS, Jan. 2017.

[96] F. Dernoncourt, J. Y. Lee, and P. Szolovits, "Neuroner: an easy-to-use program for named-entity recognition based on neural networks," in Proceedings of the 2017 Conference on Empirical Methods in Natural Language Processing: System Demonstrations. Association for Computational Linguistics,
pp. 97–102, 2017.

[97] O. Uzuner, B. R. South, S. Shen, and S. L. DuVall, "2010 i2b2 challenge on concepts, assertions, and relations in clinical text," Journal of the American Medical Informatics Association, vol. 18, no. 5, pp. 552–556, 2011.

[98] J. L. Elman, "Finding structure in time," Cognitive science, vol. 14, no. 2, pp. 179–211, 1990.

[99] M. I. Jordan, "Serial order: A parallel distributed processing approach," in Advances in psychology. Elsevier, vol. 121, pp. 471–495, 1997.

[100] A. Stubbs, C. Kotfila, and O. Uzuner, "Automated systems for the ¨ de-identification of longitudinal clinical narratives: Overview of 2014 i2b2 health shared task track 1," Journal of biomedical informatics, vol. 58, pp. S11–S19, 2015.

[101] X. Ma and E. Hovy, "End-to-end sequence labeling via bi-directional lstm cnns-crf," in Proceedings of the 54th Annual Meeting of the Association for Computational Linguistics. Association for Computational Linguistics, pp. 1064–1074, 2016.

[102] S. G. Finlayson, P. LePendu, and N. H. Shah, "Building the graph of medicine from millions of clinical narratives," Scientific data, vol. 1, p. 140032, 2014.

[103] A. R. Aronson and F.-M. Lang, "An overview of metamap: historical perspective and recent advances," Journal of the American Medical Informatics Association, vol. 17, no. 3, pp. 229–236, 2010.

[104] O. Levy and Y. Goldberg, "Dependency-based word embeddings," in Proceedings of the 52nd Annual Meeting of the Association for Computational Linguistics, vol. 2, pp. 302–308, 2014.

[105] J. Nam, E. L. Menc´ıa, and J. F¨urnkranz, "All-in text: Learning document, label, and word representations jointly," in Thirtieth AAAI Conference on Artificial Intelligence, pp. 1948–1954, 2016.

[106] L. v. d. Maaten and G. Hinton, "Visualizing data using t-sne," Journal of machine learning research, vol. 9, pp. 2579–2605, 2008.

[107] J. Josse and S. Holmes, "Measures of dependence between random vectors and tests of independence. literature review," arXiv preprint arXiv:1307.7383, 2013.

[108] G. J. Sz´ekely, M. L. Rizzo, N. K. Bakirov et al., "Measuring and testing dependence by correlation of distances," The annals of statistics, vol. 35, no. 6, pp. 2769–2794, 2007.

[109] E. K. Proctor, J. Landsverk, G. Aarons, D. Chambers, C. Glisson, and B. Mittman, "Implementation research in mental health services: an emerging science with conceptual, methodological, and training challenges," Administration and Policy in Mental Health and Mental Health Services Research, vol. 36, no. 1, pp. 24–34, 2009.

[110] X. Rong, "word2vec parameter learning explained," arXiv preprint arXiv:1411.2738, 2014.

[111] J. Niu, Y. Yang, S. Zhang, Z. Sun, and W. Zhang, "Multi-task character level attentional networks for medical concept normalization," Neural Processing Letters, pp. 1–18, 2018.

[112] Z. Liu, M. Yang, X. Wang, Q. Chen, B. Tang, Z. Wang, and H. Xu, "Entity recognition from clinical texts via recurrent neural network," BMC Medical Informatics and Decision Making, vol. 17, no. 2, p. 67, 2017.

[113] N. Limsopatham and N. Collier, "Adapting phrase-based machine translation to normalise medical terms in social media messages," in Proceedings of the 2015 Conference on Empirical Methods in Natural Language Processing. Lisbon, Portugal: Association for Computational Linguistics, pp. 1675–1680, 2015.

[114] J. Y. Lee, F. Dernoncourt, O. Uzuner, and P. Szolovits, "Feature augmented neural networks for patient note de-identification," in Proceedings of the Clinical Natural Language Processing Workshop (ClinicalNLP). Osaka, Japan: The COLING 2016 Organizing Committee, pp. 17–22, 2016.

[115] J. Xie, X. Liu, and D. Dajun Zeng, "Mining e-cigarette adverse events in social media using bi-lstm recurrent neural network with word embedding representation," Journal of the American Medical Informatics Association, vol. 25, no. 1, pp. 72–80, 2017.

[116] R. Chalapathy, E. Zare Borzeshi, and M. Piccardi, "Bidirectional lstm crf for clinical concept extraction," in Proceedings of the Clinical Natural Language Processing Workshop (ClinicalNLP). The COLING 2016 Organizing Committee, pp. 7–12, 2016.

[117] R. Chalapathy, E. Z. Borzeshi, and M. Piccardi, "An investigation of recurrent neural architectures for drug name recognition," arXiv preprint arXiv:1609.07585, 2016.

[118] I. Banerjee, M. C. Chen, M. P. Lungren, and D. L. Rubin, "Radiology report annotation using intelligent word embeddings: Applied to multiinstitutional chest ct cohort," Journal of biomedical informatics, vol. 77, pp. 11–20, 2018.

[119] I. J. Unanue, E. Z. Borzeshi, and M. Piccardi, "Recurrent neural networks with specialized word embeddings for health-domain named-entity recognition," Journal of biomedical informatics, vol. 76, pp. 102–109, 2017.

[120] A. Cocos, A. G. Fiks, and A. J. Masino, "Deep learning for pharmacovigilance: recurrent neural network architectures for labeling adverse drug reactions in twitter posts," Journal of the American Medical Informatics Association, vol. 24, no. 4, pp. 813–821, 2017.

[121] S. Henry, C. Cuffy, and B. T. McInnes, "Vector representations of multiword terms for semantic relatedness," Journal of biomedical informatics, vol. 77, pp. 111–119, 2018.

[122] H. Zhu, I. C. Paschalidis, and A. Tahmasebi, "Clinical concept extraction with contextual word embedding," arXiv preprint arXiv:1810.10566, 2018.

[123] M. Peters, M. Neumann, M. Iyyer, M. Gardner, C. Clark, K. Lee, and L. Zettlemoyer, "Deep contextualized word representations," in Proceedings of the 2018 Conference of the North American Chapter of the Association for Computational Linguistics: Human Language Technologies. Association for Computational Linguistics, pp. 2227–2237, 2018.

[124] A. Hliaoutakis, "Semantic similarity measures in mesh ontology and their application to information retrieval on medline,"Master's thesis, 2005.

[125] B. McCann, J. Bradbury, C. Xiong, and R. Socher, "Learned in translation: Contextualized word vectors," in Advances in Neural Information Processing Systems, pp. 6294–6305, 2017.

[126] Y. Tao, B. Godefroy, G. Genthial, and C. Potts, "Effective feature representation for clinical text concept extraction," arXiv preprint arXiv:1811.00070, 2018.

[127] O. Melamud, J. Goldberger, and I. Dagan, "context2vec: Learning generic context embedding with bidirectional lstm," in Proceedings of The 20th SIGNLL Conference on Computational Natural Language Learning. Association for Computational Linguistics, pp. 51–61, 2016.

[128] M. Peters, W. Ammar, C. Bhagavatula, and R. Power, "Semi-supervised sequence tagging with bidirectional language models," in Proceedings of the 55th Annual Meeting of the Association for Computational Linguistics. Association for Computational Linguistics, pp. 1756–1765, 2017.

[129] M. Baroni, G. Dinu, and G. Kruszewski, "Don't count, predict! a systematic comparison of context-counting vs. context-predicting semanticvectors," inProceedings of the 52nd Annual Meeting of the Association for Computational Linguistics.Association for Computational Linguis-tics, pp. 238–247,2014.

[130] T. Mikolov, I. Sutskever, K. Chen, G. Corrado, and J. Dean, "Distributed representations of words and phrases and their compositionality," inPro-ceedings of the 26th International Conference on Neural Information Pro-cessing Systems - Volume 2, ser. NIPS'13 , pp. 3111–3119, 2013.

[131] S. Pakhomov, B. McInnes, T. Adam, Y. Liu, T. Pedersen, and G. B.Melton, "Semantic similarity and relatedness between clinical terms: anexperimental study," inAMIA annual symposium proceedings, vol. 2010.American Medical Informatics Association, p. 572,2010.

[132] S. V. Pakhomov, T. Pedersen, B. McInnes, G. B. Melton, A. Ruggieri, andC. G. Chute, "Towards a framework for developing semantic relatednessreference standards,"Journal of biomedical informatics, vol. 44, no. 2, pp.251–265, 2011.

[133] T. Pedersen, S. V. Pakhomov, S. Patwardhan, and C. G. Chute, "Measures of semantic similarity and relatedness in the biomedical domain,"Journalof biomedical informatics, vol. 40, no. 3, pp. 288–299, 2007.